\documentclass[letterpaper, 10 pt, conference]{ieeeconf}  

\IEEEoverridecommandlockouts                              

\usepackage{graphics} 
\usepackage{epsfig} 
\usepackage{mathptmx} 
\usepackage{times} 
\usepackage{amsmath} 
\usepackage{amssymb}  

\usepackage{hyperref}
\title{\LARGE \bf
Walking on Partial Footholds Including Line Contacts with the Humanoid Robot Atlas*
}

\author{Georg Wiedebach$^{1}$, Sylvain Bertrand$^{1}$, Tingfan Wu$^{1}$, Luca Fiorio$^{2}$, Stephen McCrory$^{1}$, Robert Griffin$^{1}$,\\
        Francesco Nori$^{2}$, and Jerry Pratt$^{1}$
\thanks{*This work was funded through the National Robotics Initiative by the National Aeronautics and Space Administration (grant number NNX12AP97G).}
\thanks{$^{1}$The author is with the Florida Institute for Human and Machine Cognition,
        40 S Alcaniz St, Pensacola, FL 32502, United States}%
\thanks{$^{2}$The author is with the iCub Facility Department, Istituto Italiano di Tecnologia,
        Via Morego 30, Genova, Italy.}%
}

\begin{document}

\maketitle
\thispagestyle{empty}
\pagestyle{empty}

\begin{abstract}
We present a method for humanoid robot walking on partial footholds such as small stepping stones and rocks with sharp surfaces. Our algorithm does not rely on prior knowledge of the foothold, but information about an expected foothold can be used to improve the stepping performance. After a step is taken, the robot explores the new contact surface by attempting to shift the center of pressure around the foot. The available foothold is inferred by the way in which the foot rotates about contact edges and/or by the achieved center of pressure locations on the foot during exploration. This estimated contact area is then used by a whole body momentum-based control algorithm. To walk and balance on partial footholds, we combine fast, dynamic stepping with the use of upper body angular momentum to regain balance. We applied this method to the Atlas humanoid designed by Boston Dynamics to walk over small contact surfaces, such as line and point contacts. We present experimental results and discuss performance limitations.
\end{abstract}

\section{INTRODUCTION}

Humans are highly mobile and can walk over pointy rocks with the ground only partially supporting their feet. They walk on edges and point contacts without much trouble balancing. To achieve this they use a combination of various balancing mechanisms such as fast and dynamic stepping, shifting their Center of Pressure (CoP) within the available foothold (often referred to as the ``ankle strategy''), and angular momentum-based methods such as ``lunging'' their upper body (often referred to as the ``hip strategy'') \cite{horak1986central} or moving their arms \cite{pijnappels2010armed}.

\begin{figure}
  \centering
  \raisebox{-0.5\height}{%
    \includegraphics[width=0.9\columnwidth]{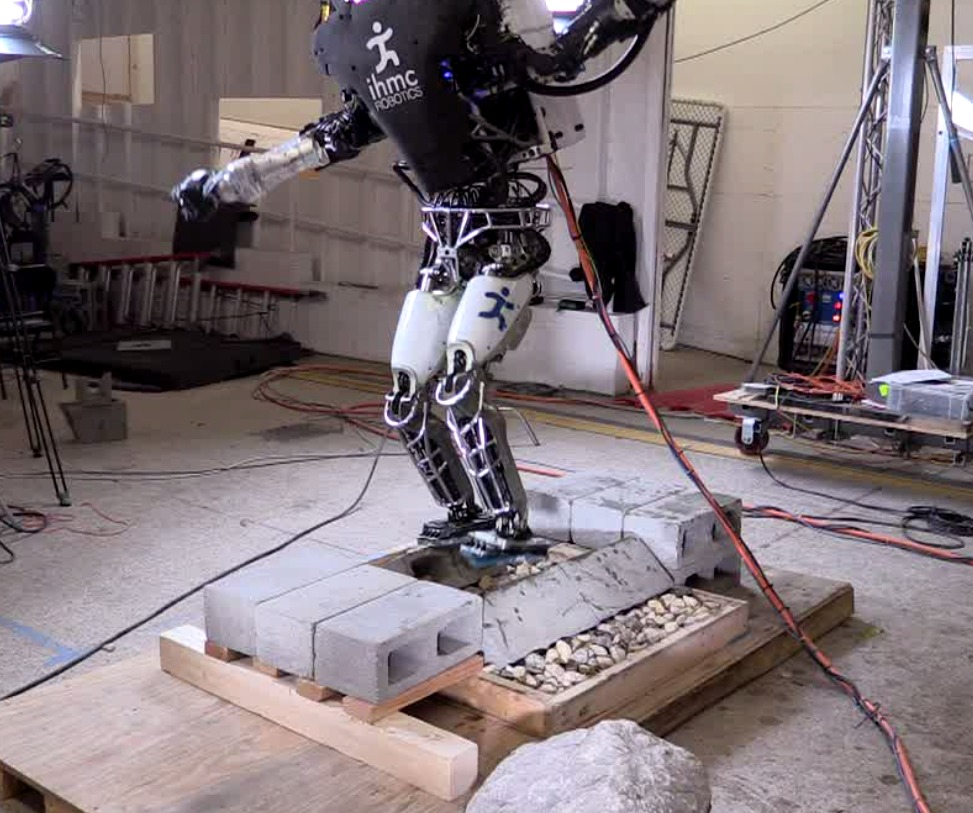}
  }
  \caption{Atlas walking on straight line contacts. Both cinder block rows are tilted and provide only line support for the feet of the robot. We present the approach taken to achieve the walking behavior seen here.}
  \label{fig:straight_lines}
\end{figure}

Humanoid robots can use these same strategies as humans, to varying effect \cite{stephens2007humanoid}. Humanoid robots that take steps can maintain balance through foot placement. Those with finite-sized feet and controllability of their CoP can shift the CoP around in their base of support. Those with good control of their upper body can use angular momentum to help maintain balance. There has been theoretical work on using a combination of multiple recovery strategies \cite{aftab2012ankle,atkeson2007multiple} in simulation to recover from disturbances such as pushes. When walking on unknown and potentially small footholds, the choice of balance strategy is essential. Taking a step to recover balance might not be possible if the potential footholds are sparse. CoP strategies might be limited by small support polygons. Angular momentum-based balancing as seen in Fig. \ref{fig:straight_lines} is generally available, but should be kept minimal since its effectiveness is limited by the robot's kinematics. In addition, any lunging maneuver needs to be reverted when the robot brings its torso back into an upright position.

A second challenge is the estimation of the current foothold. Elevation maps of the environment built from 3d sensors, such as Lidars, can be used to find possible stepping regions \cite{nishiwaki2012autonomous} or to obtain a guess about the upcoming foothold. However, when walking on footholds with a small size this method might not offer the required precision. In some cases, such as walking on terrain that is covered in leaves or mud, visual foothold estimation can be misleading. To be useful in real world applications, a robot should be able to deal with unexpected partial footholds. Much like a human with closed eyes, it should carefully step and feel around with its foot before shifting its weight fully to the newly gained foothold. Humans have dense pressure sensing on the bottom of their feet, which allows them to instantly feel an ``image'' of the ground in fairly high resolution. Most state of the art humanoid robots, however, only have force sensors that allow them to detect the ground wrench, allowing for estimation of the immediate CoP, but not the full contact area. Instead, the available contacts must be inferred by the dynamic interaction of the foot with the ground, for example by observing rotation about contact edges and/or by examining achieved CoP locations during active exploration of the contact surface.

\begin{figure}
  \centering
  \includegraphics[width=\columnwidth]{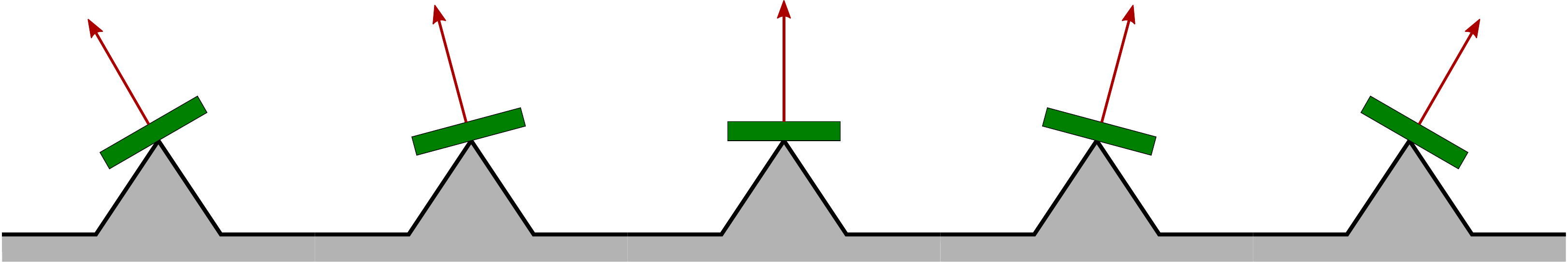}
  \caption{When walking on sharp surfaces, we keep the foot in a close to flat orientation instead of having it conform to the ground. When balancing on a line, the foot orientation parallel to the line is held using position control while the torque perpendicular to the line is used to shift the CoP along the line contact. By rotating the foot, the effective surface normal can be altered to prevent slipping. If the foot surface is compliant, it can grip the sharp surface, further preventing slip.}
  \label{fig:line_contact_normal}
\end{figure}

Most humanoid walking approaches in the literature assume full footholds and focus on footstep planning on potentially slanted but locally flat surfaces with large contact areas \cite{deits2014footstep,stumpf2014supervised,nishiwaki2014planning}. An example of this is the recent DARPA Robotics Challenge, where robots had to traverse a field of slanted cinder blocks. Other references, such as \cite{manchester2009stable,zucker2011optimization} demonstrate the ability to traverse jagged rocks using robots with point feet. To our knowledge there has been little work on the task of walking over pointy rocks using bipeds with feet. When walking over rubble and small debris, torque controlled robots can rely on ankle compliance to let their foot conform to the ground. Here we focus on situations where this approach is not feasible when the available contact is severely limited. In this case, the foot should not conform to the ground but rather stay close to a flat orientation to avoid reaching ankle joint limits. Fig. \ref{fig:line_contact_normal} illustrates this situation with a front view of walking over a line or point contact. One advantage of walking on such surfaces is that the effective friction cone of the ground can be made quite large since the humanoid can angle its foot to achieve a large range of effective surface normals. Orienting the foot so that the surface normal is favorable to the desired ground reaction forces can significantly reduce the probability of slipping as compared to walking on flat but tilted surfaces.

To generate whole body motions for balancing and foothold exploration, we use the momentum-based control framework described in \cite{koolen2016design}, which was first introduced in \cite{orin2013centroidal}. Since then, optimization based control frameworks became increasingly popular \cite{feng2015optimization,kuindersma2014efficiently}. We demonstrate the flexibility and the capabilities of the control algorithm. We show that it can handle non-trivial foothold geometries, foot exploration, online changes of the footholds, and complex balancing motions.

In our approach we combine fast swing times with upper body angular momentum to walk and balance on small footholds. After each step, we pause and explore the newly gained foothold before continuing to walk. First, we give an overview of the control framework used throughout this work in Section \ref{sec:control_framework}. The estimation of the support area is described in Section \ref{sec:foothold_detection}. The stepping and balancing is the subject of Section \ref{sec:walking_and_balance_control}. Finally, we present experiments done with the Atlas robot in Section \ref{sec:experiments} that show the feasibility and limitations of the proposed method.

\section{CONTROL FRAMEWORK}
\label{sec:control_framework}

A momentum-based control framework is used to control the humanoid Atlas and is introduced in \cite{koolen2016design}. It uses a Quadratic Program (QP) to optimize a cost function at every controller time step. The core of the controller has been improved to enable the robot to explore footholds and balance with a limited support area. The QP formulation has been extended as follows:
\begin{equation}
\begin{aligned}
  & \underset{\dot{\mathbf{v}}_d, \mathbf{\rho}}{\text{min}} & & c_{\dot{\mathbf{h}}_d} + c_\mathbf{J} + c_\mathbf{P} + c_{\mathbf{\rho}} + c_{\dot{\mathbf{v}}_d}                     \\
  & \text{s.t.}                            & & \mathbf{A}\dot{\mathbf{v}}_d + \dot{\mathbf{A}}\mathbf{v} = \mathbf{W}_g + \mathbf{Q}_{CoM}\mathbf{\rho} + \sum\nolimits_i \mathbf{W}_{ext,i} \\
  &                                        & & \mathbf{\rho}_{min} \leq \mathbf{\rho}                                                                                                  \\
  &                                        & & \dot{\mathbf{v}}_{min} \leq \dot{\mathbf{v}}_d \leq \dot{\mathbf{v}}_{max}
\end{aligned}
\end{equation}
The terms of the objective function are defined as
\begin{equation*}
\begin{aligned}
  & \text{Momentum Objective:}      & & c_{\dot{\mathbf{h}}_d} = \left(\mathbf{A}\dot{\mathbf{v}}_d - \mathbf{b}\right)^T \cdot \mathbf{C}_{\dot{\mathbf{h}}}   \cdot \left(\mathbf{A}\dot{\mathbf{v}}_d - \mathbf{b}\right) \\
  & \text{Motion Objective:}        & & c_\mathbf{J}           = \left(\mathbf{J}\dot{\mathbf{v}}_d - \mathbf{p}\right)^T \cdot \mathbf{C}_\mathbf{J}           \cdot \left(\mathbf{J}\dot{\mathbf{v}}_d - \mathbf{p}\right) \\
  & \text{Contact Force Objective:} & & c_\mathbf{P}           = \left(\mathbf{P}\mathbf{\rho}      - \mathbf{r}\right)^T \cdot \mathbf{C}_\mathbf{P}           \cdot \left(\mathbf{P}\mathbf{\rho}      - \mathbf{r}\right) \\
  & \text{Contact Force Cost:}      & & c_{\mathbf{\rho}}      = \mathbf{\rho}^T                                          \cdot \mathbf{C}_{\mathbf{\rho}}      \cdot \mathbf{\rho}                                          \\
  & \text{Joint Acceleration Cost:} & & c_{\dot{\mathbf{v}}_d} = \dot{\mathbf{v}}_d^T                                     \cdot \mathbf{C}_{\dot{\mathbf{v}}_d} \cdot \dot{\mathbf{v}}_d
\end{aligned}
\end{equation*}
where:
\begin{itemize}
 \item $\dot{\mathbf{v}}_d$ are the desired joint accelerations and $\mathbf{\rho}$ consists of the contact point force magnitudes as introduced in \cite{koolen2016design}.
 \item $\mathbf{A}$ is the centroidal momentum matrix and $\mathbf{b} = \dot{\mathbf{h}}_d - \dot{\mathbf{A}}\mathbf{v}$ with $\dot{\mathbf{h}}_d$ being the desired rate of change of linear momentum and $\mathbf{v}$ denoting the joint velocities. The momentum objective only incorporates the linear part of the centroidal momentum. This leaves the angular momentum free to be determined by the optimization.
 \item $\mathbf{J} = \left[ \mathbf{J}_1^T \dots \mathbf{J}_k^T \right]^T$ and $\mathbf{p} = \left[ \mathbf{p}_1^T \dots \mathbf{p}_k^T \right]^T$ are the concatenated Jacobian matrices and objectives for $k$ desired motions.
 \item $\mathbf{P}$ and $\mathbf{r}$ can be used to define objectives on the ground reaction forces and will be defined later in this section.
 \item $\mathbf{Q}_{CoM}$ is the Jacobian matrix from the contact force space $\mathbf{\rho}$ to the centroidal frame.
 \item $\mathbf{W}_{ext,i}$ are $i$ external wrenches on the robot and $\mathbf{W}_g$ is the gravitational wrench.
 \item $\mathbf{\rho}_{min}$ is the lower bound on $\mathbf{\rho}$, used to enforce contact unilaterality.
 \item $\dot{\mathbf{v}}_{min}$ and $\dot{\mathbf{v}}_{max}$ are bounds on the joint acceleration, used to enforce joint angle limits. We compute the acceleration bounds every tick based on the current joint position and velocity to avoid hitting position limits.
 \item $\mathbf{C}_{\dot{\mathbf{h}}}$, $\mathbf{C}_\mathbf{J}$, $\mathbf{C}_\mathbf{P}$, $\mathbf{C}_{\mathbf{\rho}}$, and $\mathbf{C}_{\dot{\mathbf{v}}_d}$ are positive definite cost function weighting matrices.
\end{itemize}

The first main contribution of this work is that all the desired motions are now part of the objective function, whereas they were previously formulated as equality constraints. The biggest difference we observe is that the momentum objective used to maintain balance can take priority over the motion objectives. We have introduced and tuned weights $\mathbf{C}$ for all terms of the objective function. The weight magnitudes can be adjusted by the walking control algorithm and may be changed online according to the current state of the robot. This enables the optimization to prioritize some objectives over others. For example, during stable stance on both feet, a high weight can be assigned to manipulation tasks to achieve good end effector trajectory tracking. On the other hand, during difficult balancing maneuvers, the weight on the momentum objective is increased, sacrificing precise upper body motions to help regain balance.

The second contribution is the introduction of an objective for the ground reaction force with the $c_\mathbf{P}$ cost term. It can be used to control the desired CoP of an end effector in contact with the environment by favoring a certain distribution in the ground reaction force magnitudes. For a single foot $f$ in contact with the environment, the position of the local CoP on the sole can be written as
\begin{equation}
  \mathbf{x}_{CoP, f} = \mathbf{P}_f \cdot \mathbf{\rho}_f \text{ ,}
\end{equation}
\begin{equation}
  \mathbf{P}_f \equiv \frac{1}{F_f^Z} \cdot \mathbf{S} \cdot \mathbf{Q}_f \text{ ,}
\end{equation}
\begin{equation}
  \mathbf{S} = \begin{bmatrix} 0 & -1 & 0 & 0 & 0 & 0 \\ 
                               1 &  0 & 0 & 0 & 0 & 0 \end{bmatrix} \text{ .}
\end{equation}
Here, $\mathbf{Q}_f$ transforms all the force magnitudes $\mathbf{\rho}_{f}$ of the foot contact points to a single wrench at the contact plane origin. The matrix $\mathbf{S}$ is a selection matrix selecting only the horizontal torques induced by the ground contact forces at the contact plane frame which is aligned with the bottom of the foot. The value of $F_f^Z$ is the vertical foot force magnitude computed during the previous control step. The full $\mathbf{P}$ matrix in the objective can then be assembled from all $m$ flat end effectors with a CoP position objective. The vector $\mathbf{r}$ contains all desired CoP positions $\mathbf{x}_{CoP,f,d}$:
\begin{equation}
  \mathbf{P} = diag \left( \mathbf{P}_1 \dots \mathbf{P}_m \right) \text{ ,}
\end{equation}
\begin{equation}
  \mathbf{r} = \left[ \mathbf{x}_{CoP,1,d}^T \dots \mathbf{x}_{CoP,m,d}^T \right]^T \text{ .}
\end{equation}

\section{FOOTHOLD DETECTION}
\label{sec:foothold_detection}

\begin{figure}
  \centering
  \includegraphics[width=\columnwidth]{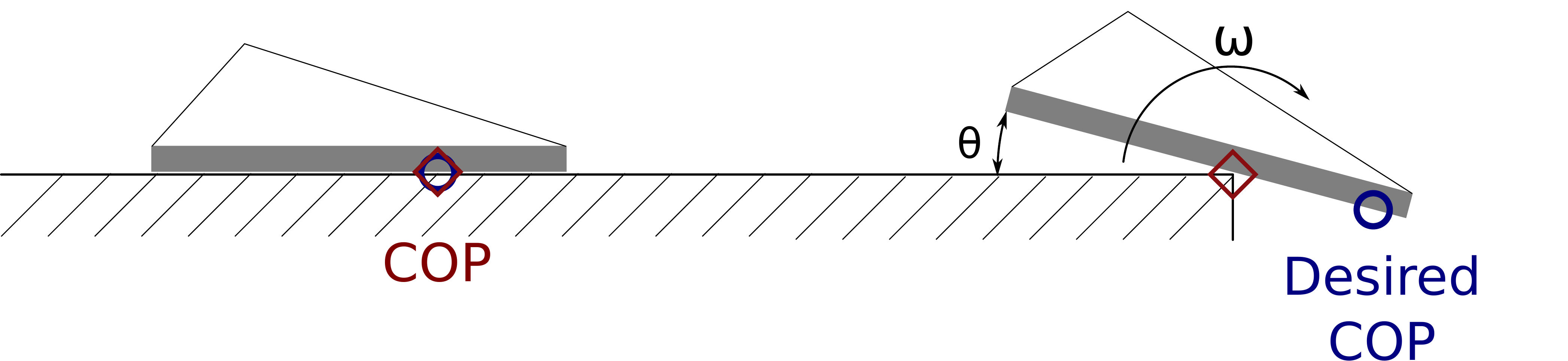}
  \caption{Foothold Exploration. Side view of two feet with one foot rotating about the edge of the supporting ground. If foot rotation is detected, the control algorithm adjusts the foothold and removes the part that cannot support weight. The rotation speed $\omega$ and the angle between foot and ground $\theta$ are estimated and serve as indicators for foot rotation.}
  \label{fig:side_view_footrotation}
\end{figure}

To balance and walk it is helpful to have an estimate of the current support polygon as the area where the CoP can be placed. While walking over uncertain terrain, the new foothold is explored after each step. If there is no additional information about the foothold available we start by assuming the full foot to be in contact with the ground. Then the robot starts to shift its local foot CoP around within the foothold. If the position of the desired CoP is in an area of the foot that is supported by the ground, the measured CoP will closely follow the desired one. However, if the desired CoP is moving outside the base of support, the foot will start to rotate about the edge of the support (Fig. \ref{fig:side_view_footrotation}). We use this rotation of the foot to refine the estimate of our support polygon by cutting out the part of the foothold that was not able to carry weight.

To detect foot rotation, we implemented two methods: one based on the measured rotational velocity of the foot and one based on the geometry of the foot and ground planes. The first method relies on accurate joint velocity measurements from the state estimator to compute the rotational speed and axis of rotation of the foot. A threshold on the rotational speed $\omega$ determines if the foothold should be adjusted. To determine the foot rotation geometrically, the foot plane is intersected with the ground plane. The intersection corresponds to the axis of the rotation and the angle $\theta$ between the planes equals the foot angle with respect to the ground. This requires an estimate of the ground plane normal and is not as fast as the velocity based detection, but is more robust to noise. Once a foot rotation is detected, the foothold is cropped at the line of rotation and the part of the foothold that would not support the desired CoP is removed. After exploration, we regulate the foot orientation to stay flat by adding an angular acceleration objective for each foot. This objective becomes active when the CoP of that foot comes close to an edge of the foothold. The acceleration driving the foot to a flat orientation is computed using a PD control law.

\begin{figure}
  \centering
  \raisebox{-0.5\height}{%
    \includegraphics[width=0.41\columnwidth]{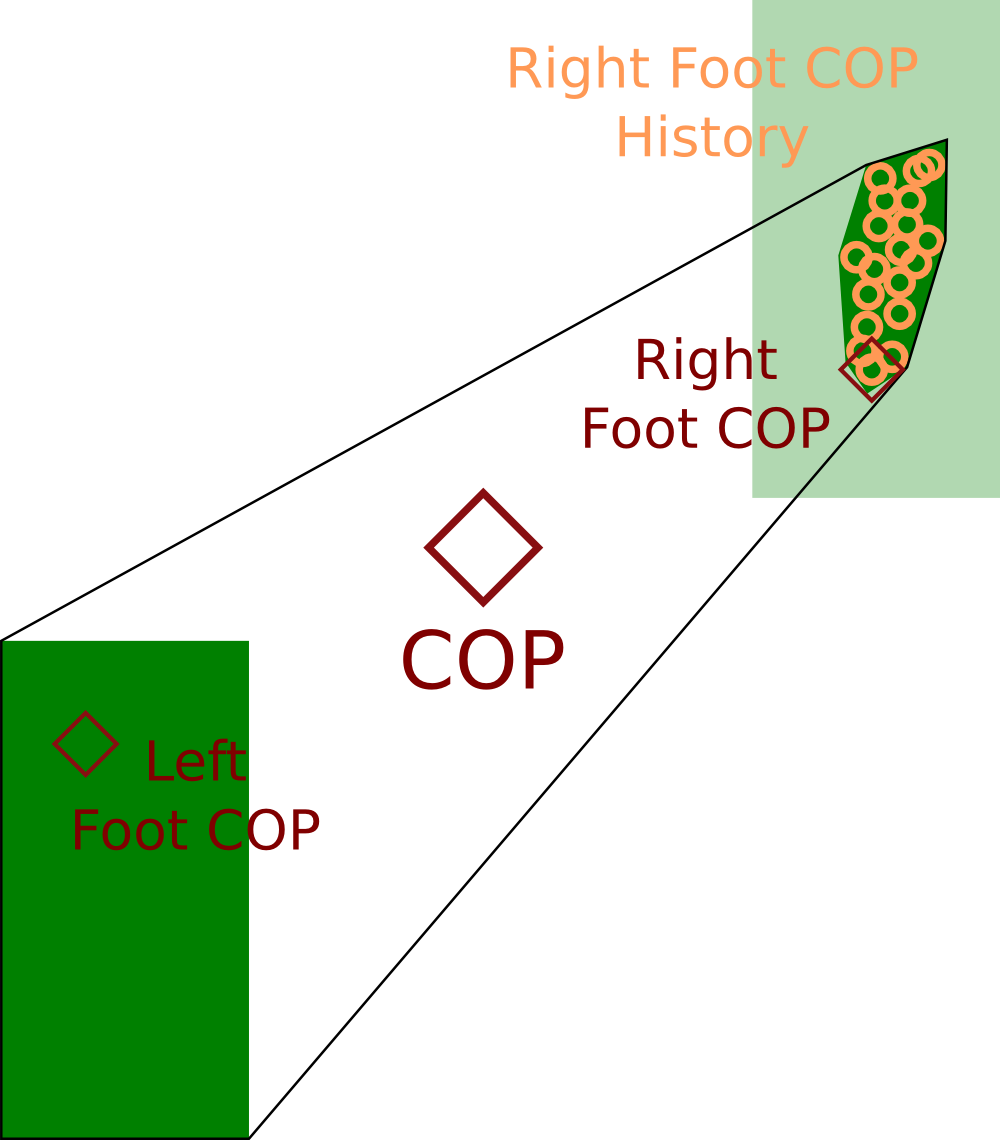}
  }
  \raisebox{-0.5\height}{%
    \includegraphics[width=0.41\columnwidth]{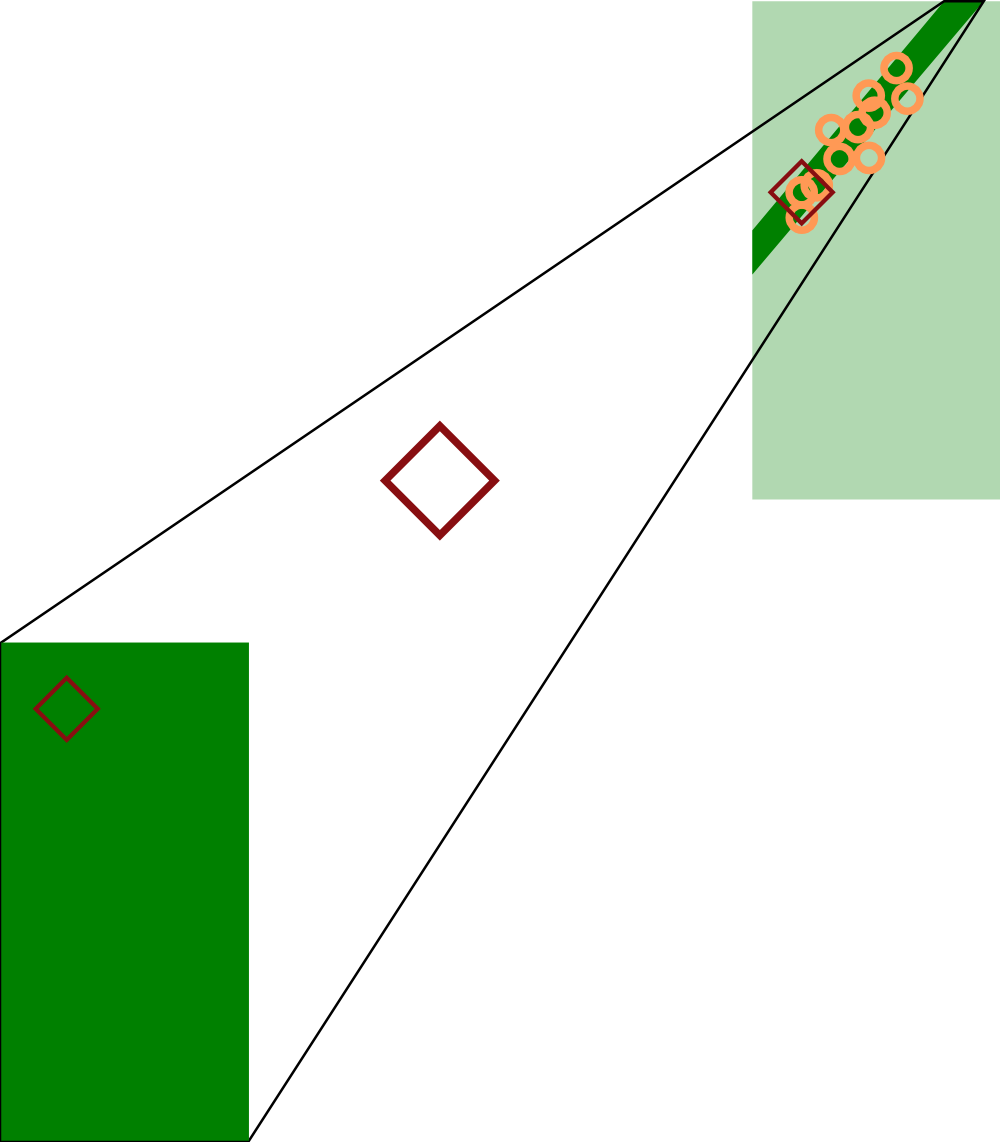}
  }
  \caption{We can use the CoP history to get a conservative estimate on the foothold by computing the convex hull of all measured CoP positions, or to fit a known terrain surface (such as a line) to the foot.}
  \label{fig:cop_fitting}
\end{figure}

These foothold detection methods provide a quick estimate of the actual foothold. To obtain a more conservative or precise estimate of the support area we can maintain the set of measured CoP positions during exploration. Since the CoP cannot leave the area of support, the convex hull of the CoP measurements is a conservative bound on the foothold. If the geometry of the foothold is given by an elevation map or by the operator, the CoP positions can be used to fit the foothold to this map of known ground contact points (see Fig. \ref{fig:cop_fitting}). For example, it might be known from the elevation map that the foot contact is a line. To find the exact foothold we then fit a line to the measured CoP positions. This will account for inaccurate stepping due to swing foot tracking errors and provide the control algorithm with a precise location of the ground contact on the foot. To account for foot slippage, new measurements are trusted more heavily than previous ones.

Each foothold is represented as a collection of contact points at the corners of the support area \cite{koolen2016design}. Although the exploration of the foothold can result in a varying number of corner points, we found that it is sufficient in practice to approximate it with a constant number of four corner points. This simplifies the optimization problem by keeping its size constant.

\section{WALKING AND BALANCE CONTROL}
\label{sec:walking_and_balance_control}

The Instantaneous Capture Point (ICP) \cite{pratt2006capture} is our main indicator of the balancing state of the robot. It is based on the Linear Inverted Pendulum Model (LIPM) \cite{kajita20013d} and is defined as
\begin{equation}
\label{eq:icp}
  \mathbf{x}_{ICP} = \mathbf{x}_{CoM} + \tfrac{1}{\omega_0} \cdot \dot{\mathbf{x}}_{CoM} \text{ ,}
\end{equation}
where $\omega_0 = \sqrt{\tfrac{g}{z}}$ is the pendulum constant. The equations presented in this section are reduced to two dimensions, since the ground reference points are defined to lie in the ground plane. The LIPM can be extended with a flywheel at the top to simulate upper body momentum and can be used to control the walking gait of a humanoid robot \cite{koolen2012capturability,pratt2012capturability}. It can be seen from this model that the Centroidal Moment Pivot (CMP) \cite{popovic2005ground} reference point controls the ICP dynamics according to
\begin{equation}
\label{eq:icp_dynamics}
  \dot{\mathbf{x}}_{ICP} = \omega_0 \cdot \left( \mathbf{x}_{ICP} - \mathbf{x}_{CMP} \right)  \text{ .}
\end{equation}

This equation states that the ICP will diverge from the CMP with increasing speed as the distance between the two gets bigger. During regular walking the torso of the robot is controlled to be upright with near zero upper body angular momentum. If there are no torques acting on the CoM of the robot the CoP coincides with the CMP \cite{goswami2004rate}. Therefore, the CoP strategy is sufficient to keep the robot balanced as long as the ICP lies inside the support area. In cases where the ICP is outside the support area, the CoP strategy is not able to keep the robot balanced. In such a context, stepping or using angular momentum is necessary. When a step can be taken to where the ICP will be at the end of the swing, the use of angular momentum is generally avoided. When stepping is not available or is insufficient, angular momentum can be used to make the CMP leave the support area temporarily, depending on the available joint torques and range of motion. This way the ICP can be driven back towards the support to regain balance. This corresponds to an upper body lunging maneuver.

In our control framework the desired CMP position is transformed to an optimization objective: the desired rate of change of the linear, horizontal momentum of the CoM determines the position of the CMP. Therefore, we can compute
\begin{equation}
  \dot{\mathbf{h}}_{d,linear} = \tfrac{m \cdot g}{z} \left( \mathbf{x}_{CoM} - \mathbf{x}_{d,CMP}\right)
\end{equation}
from (\ref{eq:icp}) and (\ref{eq:icp_dynamics}). The subscript $d$ is short for desired and $\dot{\mathbf{h}}_{d,linear}$ refers to the horizontal component of the desired rate of change in the linear centroidal momentum. This is the input to the optimization introduced in Section \ref{sec:control_framework}. The desired position of the CMP is computed according to the control law
\begin{equation}
  \mathbf{x}_{d,CMP} = \mathbf{x}_{ICP} - \tfrac{1}{\omega_0} \cdot \dot{\mathbf{x}}_{d,ICP} + k_p \cdot \left( \mathbf{x}_{ICP} - \mathbf{x}_{d,ICP} \right)
\end{equation}
similar to the approach in \cite{koolen2016design}. Desired values are defined by a reference trajectory for the ICP that is computed based on the next footstep positions \cite{englsberger2014trajectory}.

\begin{figure}
  \centering
  \includegraphics[width=\columnwidth]{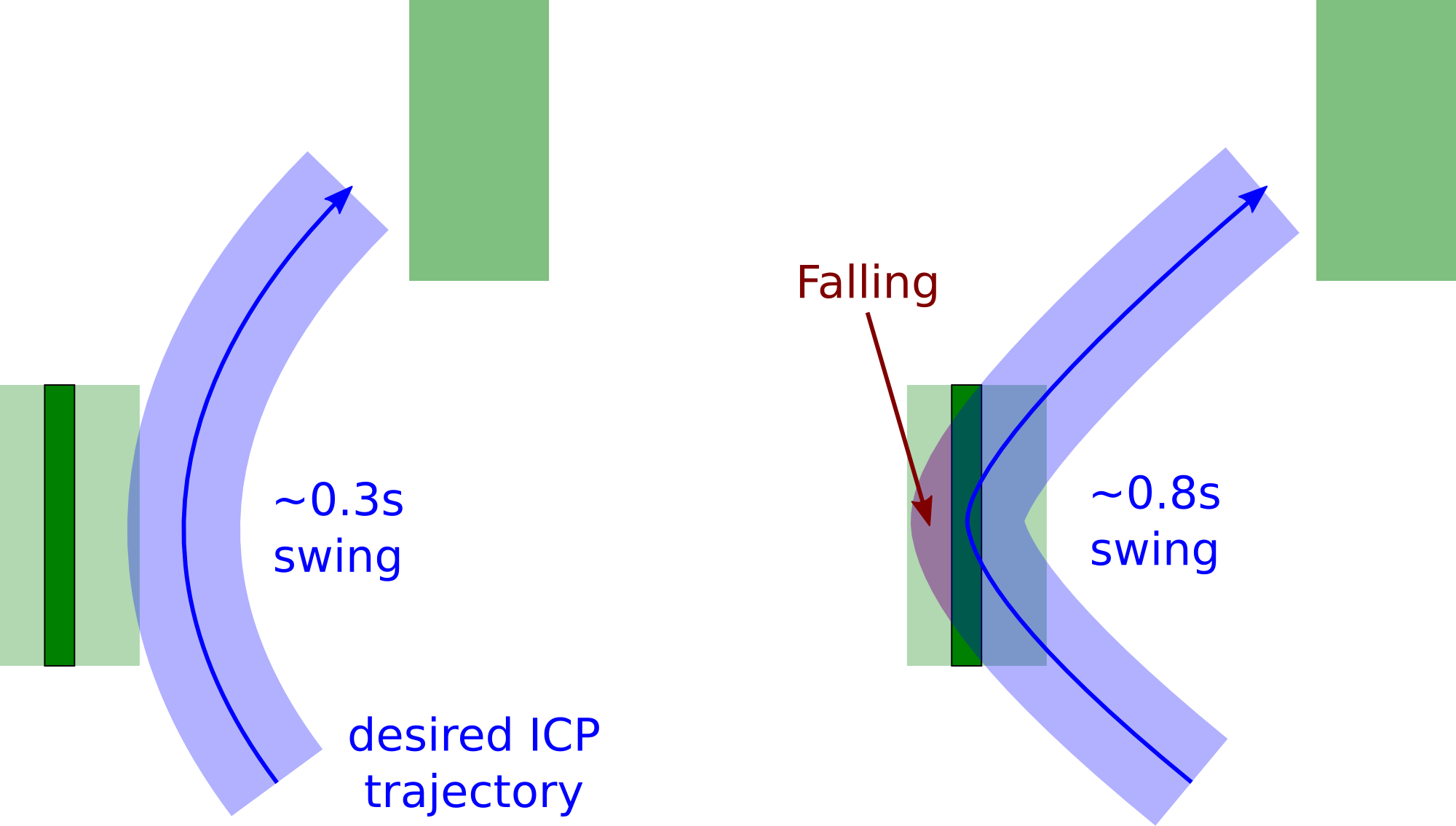}
  \caption{ICP trajectories (blue) for different swing times during a forward step with the right foot. Here the stance foot has only a line contact with the ground (dark green). The light blue area accounts for possible disturbances, as well as sensor noise and modeling uncertainty, which will move the ICP away from its nominal path. It can be seen that a fast swing time is more robust to disturbances since the ICP is less likely to exit the support area in such a way that it cannot be caught by the upcoming step (red area). When this occurs, if the footstep location cannot be significantly modified, angular momentum must be used in order to prevent a fall.}
  \label{fig:swing_time}
\end{figure}

When walking on small footholds it becomes difficult to balance on the stance foot during a step. Reducing the step time can be helpful to increase the robustness of steps on small footholds. Fig. \ref{fig:swing_time} conceptually shows a planned ICP trajectory with a bound that accounts for disturbances, sensor noise and modeling uncertainties. If the swing is fast enough, the ICP does not enter the stance foothold. Instead it gets directed towards the upcoming foothold and the robot needs to finish the step in order to regain balance. Generally it is desirable to ``fall'' towards the upcoming foothold in this way. If the stance support area is small, CoP control cannot be used to redirect the direction of the fall. The ICP control capabilities are limited by the achievable locations of the CMP. This means that a situation where the ICP leaves the support polygon, such that the robot is not falling towards the next foothold, is not recoverable unless the CMP leaves the support polygon as well.

\begin{figure}
  \centering
  \raisebox{-0.5\height}{%
    \includegraphics[height=4.8cm]{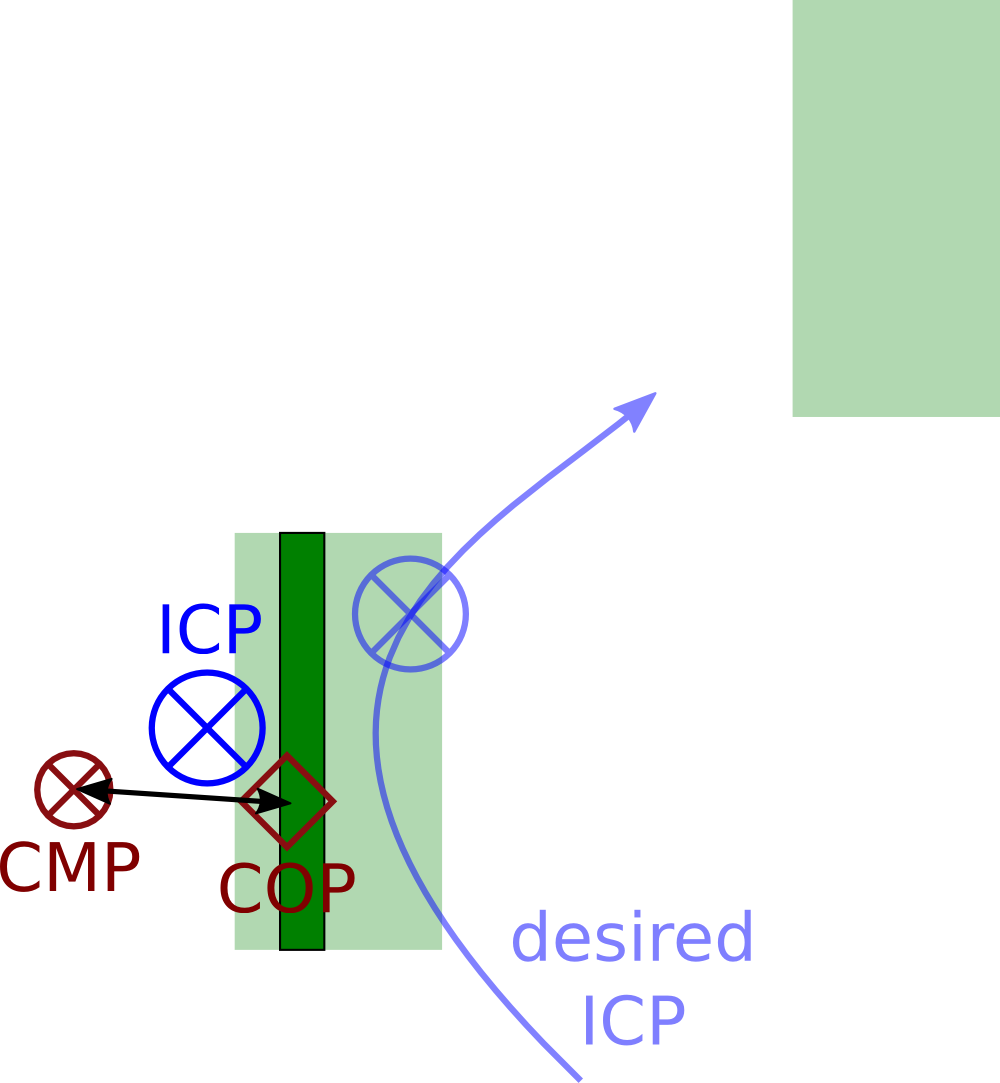}
  }
  \raisebox{-0.5\height}{%
    \includegraphics[height=4.8cm]{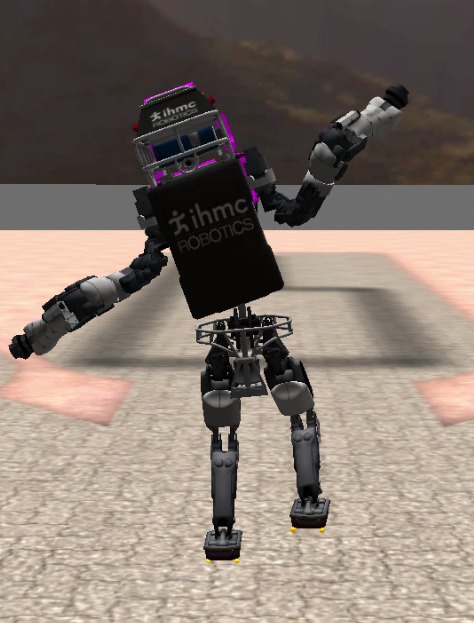}
  }
  \caption{When the ICP leaves the support area the robot can use a lunging maneuver to recover. The first picture is a top view of the ground reference points in that situation. The difference in CMP and CoP positions corresponds to a change in the angular momentum. The CMP ``pushes'' the ICP back to its desired trajectory. In the second image the simulated Atlas robot can be seen recovering from a push to the left during a step by using this lunging technique.}
  \label{fig:lunging}
\end{figure}

The difference between CoP and CMP is directly related to the change in angular momentum \cite{popovic2005ground}
\begin{equation}
  \dot{\mathbf{h}}_{angular} = \mathbf{\tau}_{CoM} = m \cdot g \cdot \left( \mathbf{x}_{CMP} - \mathbf{x}_{CoP} \right) \text{ .}
\end{equation}
In other words, torques on the CoM of the robot can be used to temporarily move the CMP outside of the support polygon of the robot and drive the ICP back inside the support area (Fig. \ref{fig:lunging}). This is necessary when the ICP leaves the support and cannot be caught by an upcoming step. Applying torques on the CoM of the robot causes lunging of the upper body and produces motions that are similar to humans recovering from strong pushes or balancing on small footholds. In our case the rate of change of angular momentum and the CoP are output of the QP and are not computed explicitly. During normal gait, the chest is kept upright limiting the generation of angular momentum. This results in the CoP staying close to the CMP. If this is not possible because the CoP can not leave the support area, the QP will be forced to use torque on the CoM to achieve its momentum objective, thereby creating a large distance between CMP and CoP. In addition we use weight scheduling on the momentum objective, increasing it when the ICP comes close to the edge of the support and decreasing it after recovering.

The lunging maneuver is limited by maximum actuator torques and joint limits of the robot, and can only be used for a short amount of time. In addition the angular momentum of the torso needs to be brought back to zero before joint limits are reached. The strategy of lunging the torso increases the robustness of walking on small footholds, particularly when swing speeds are slow, such that the robot must balance for an extended time over the support foot while waiting for the swing foot to reach the next step (Fig. \ref{fig:swing_time} right).

\begin{figure}
  \centering
  \raisebox{-0.5\height}{%
    \includegraphics[height=4.5cm]{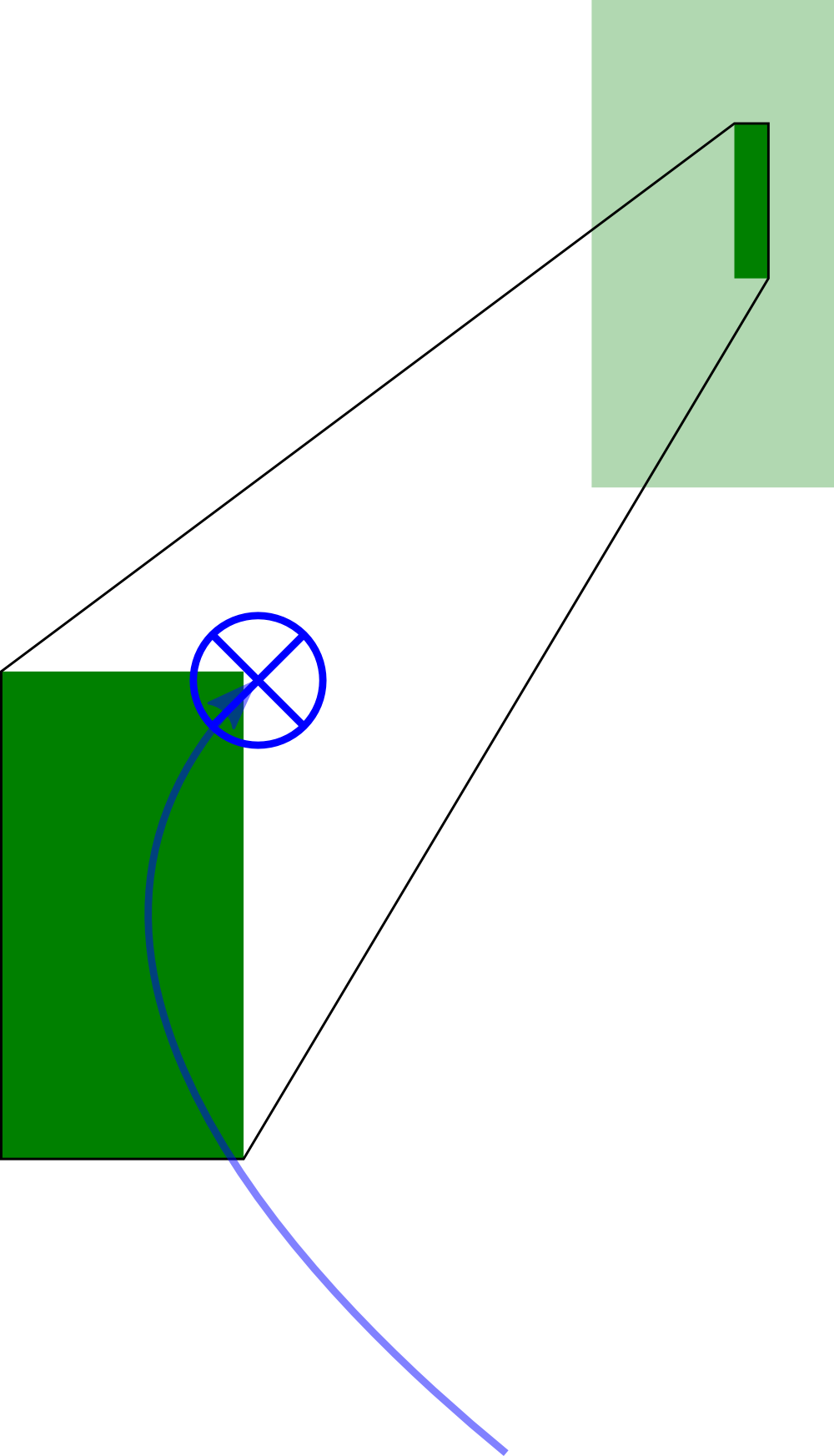}
  }\qquad
  \raisebox{-0.5\height}{%
    \includegraphics[height=4.5cm]{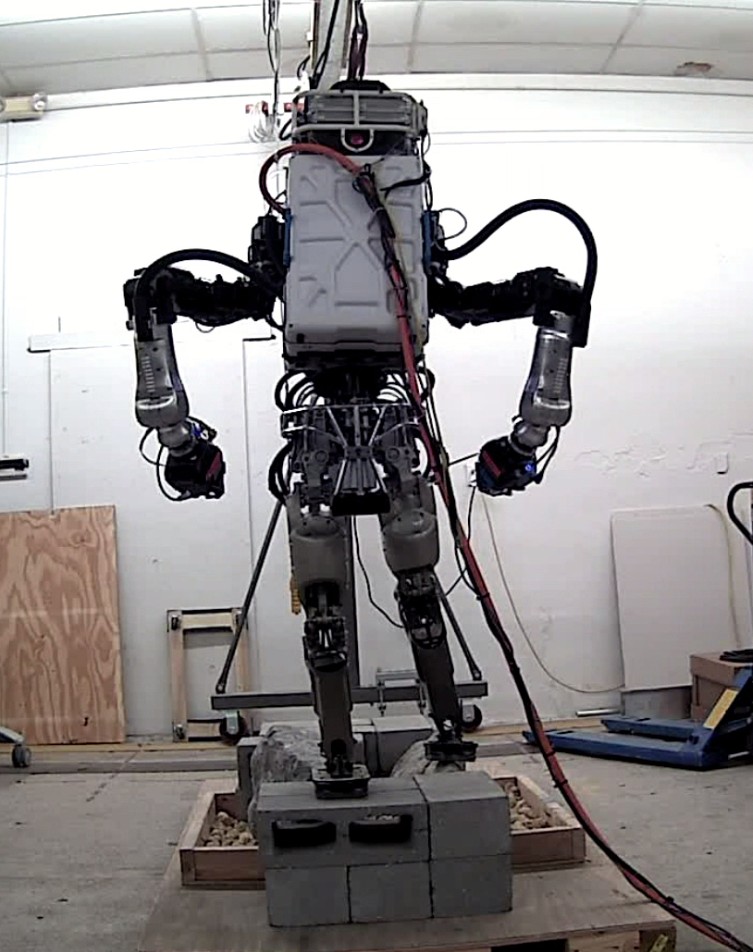}
  }
  \caption{If the expected foothold is small, the robot keeps its weight close to the stance foot and explores the upcoming foothold before shifting its weight to it. The left figures show the ICP trajectories with the final ICP at the end of the step, on the right side the Atlas robot can be seen right after taking a step that corresponds to the contact situation depicted on the left.}
  \label{fig:chicken_close}
\end{figure}

\begin{figure}
  \centering
  \raisebox{-0.5\height}{%
    \includegraphics[height=4.5cm]{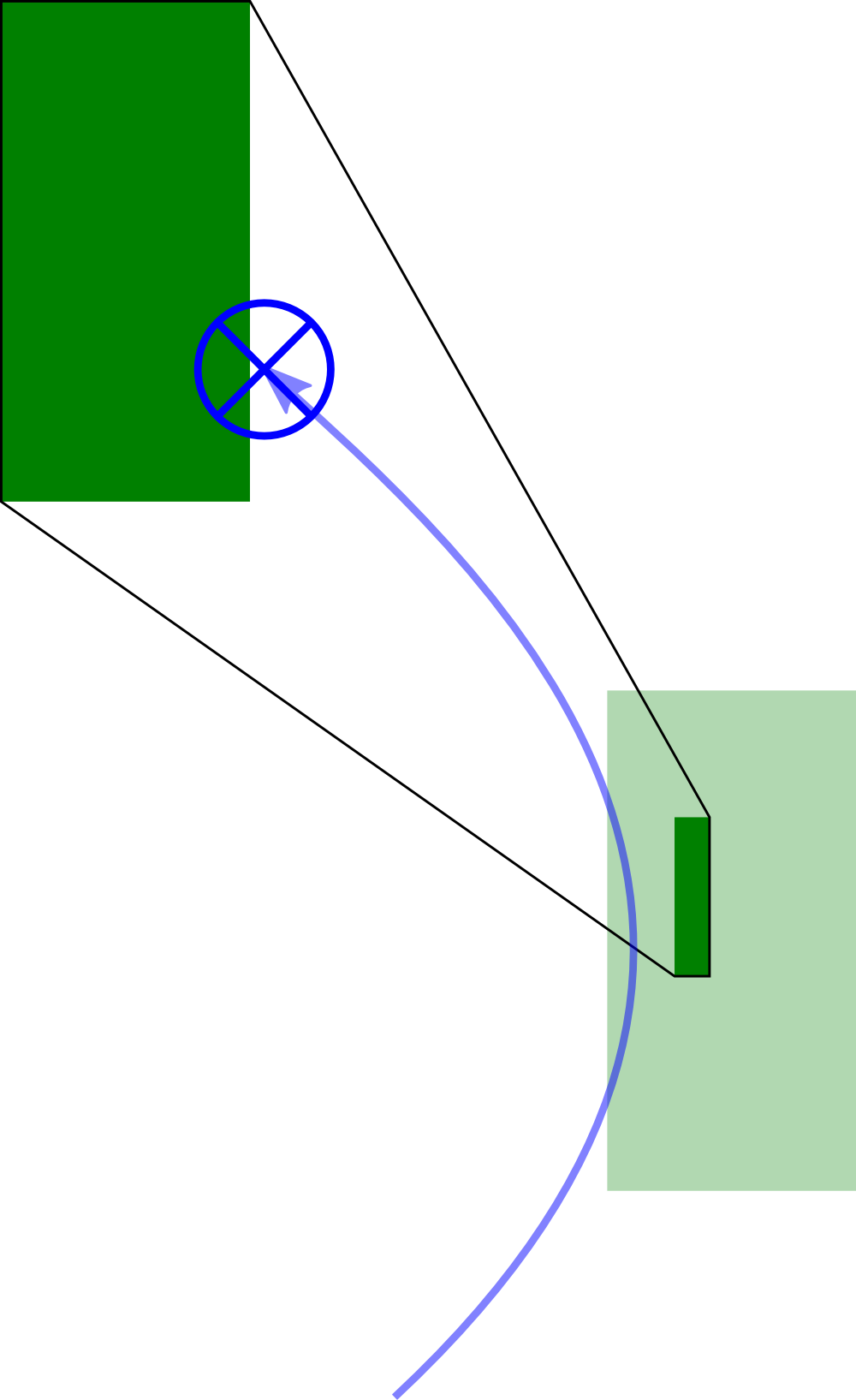}
  }\qquad
  \raisebox{-0.5\height}{%
    \includegraphics[height=4.5cm]{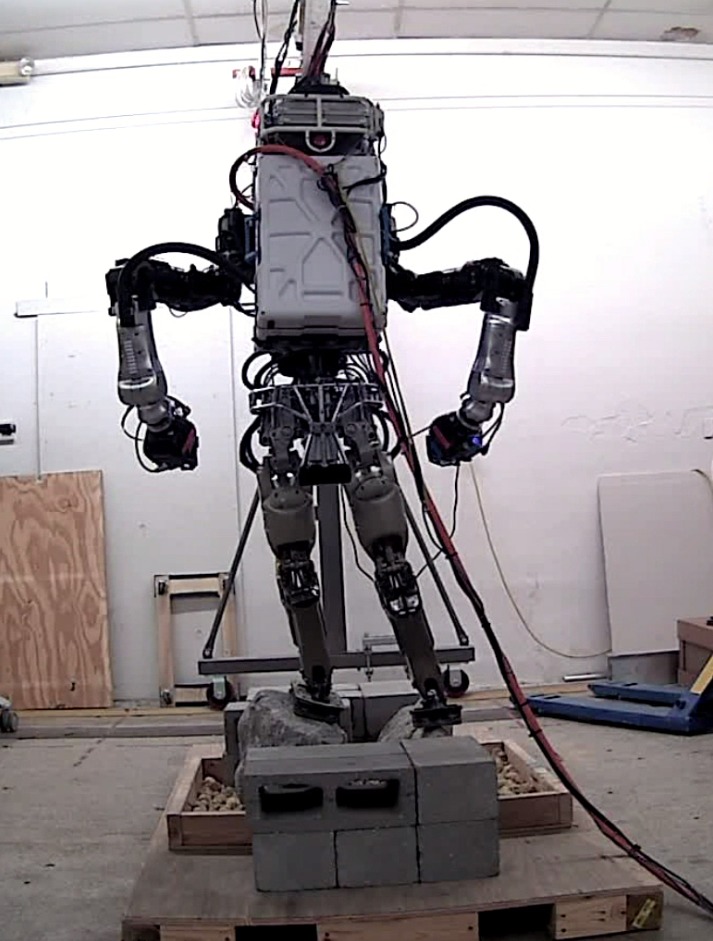}
  }
  \caption{When stepping to a good foothold while the stance foot is supported partially the robot will end the step with more of its weight resting on the good foothold.}
  \label{fig:chicken_far}
\end{figure}

To exploit knowledge of the current foothold of the robot we adjust the ICP and CoM trajectories of the robot during the step. If the current foothold is large we move the desired CoM at the end of the step closer to the current foothold. This prevents the robot from shifting a lot of weight to the new foothold. On the other hand, if the stance support area is small, we actively fall towards the upcoming foothold. In this case the desired ICP at the end of the swing is closer to the new foothold. This behavior for different foothold configurations can be seen in Fig. \ref{fig:chicken_close} and Fig. \ref{fig:chicken_far}.

\section{RESULTS \& DISCUSSION}
\label{sec:experiments}

We employed the method described above on the Atlas robot\footnote{For a summary of the results presented here refer to the video at \mbox{\url{www.youtube.com/watch?v=\_5PtxHsr038}}}. Using a combination of the foothold detection methods and balancing strategies, the robot was able to walk over different sets of limited footholds. For the results presented here a swing time of approximately 0.6\,s was used. Faster walking causes the steps to become increasingly imprecise due to larger swing foot tracking errors. The hydraulic pump of the robot additionally limits the stepping speed, as the flow rate necessary to achieve the fast leg joint motions cannot be maintained. In comparison, human step times are approximately 0.3\,s  during fast walking. Therefore, Atlas must balance longer in single support and hence resort to lunging in order to prevent falling more than a human does (see Fig. \ref{fig:swing_time}).

\begin{figure*}
  \centering
  \raisebox{-0.5\height}{%
    \includegraphics[height=5.1cm]{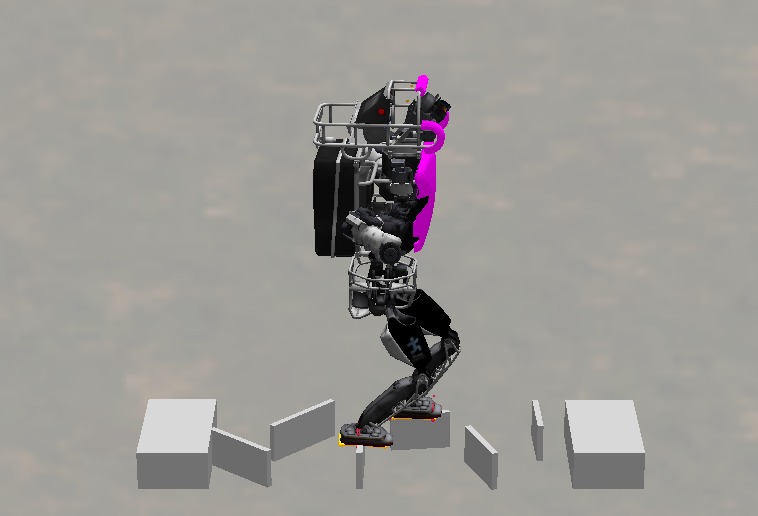}
  }
  \raisebox{-0.5\height}{%
    \includegraphics[height=5.1cm]{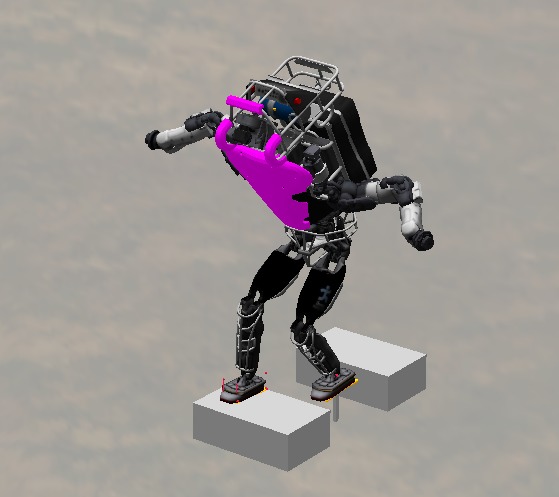}
  }
  \caption{The simulated Atlas robot walking over randomly oriented, line shaped stepping stones and a point foothold. After each step the new foothold is explored and the control algorithm adjusts the stepping accordingly.}
  \label{fig:sim_atlas}
\end{figure*}

Fig. \ref{fig:sim_atlas} shows a simulated testbed with line shaped stepping stones. After each step is taken, the robot returns to a statically stable double support stance, with the CoM inside the support polygon. The robot then explores the new foothold before continuing to walk. The exploration phase takes about 1 to 3\,s depending on the foothold.

\begin{figure}
  \centering
  \raisebox{-0.5\height}{%
    \includegraphics[width=0.85\columnwidth]{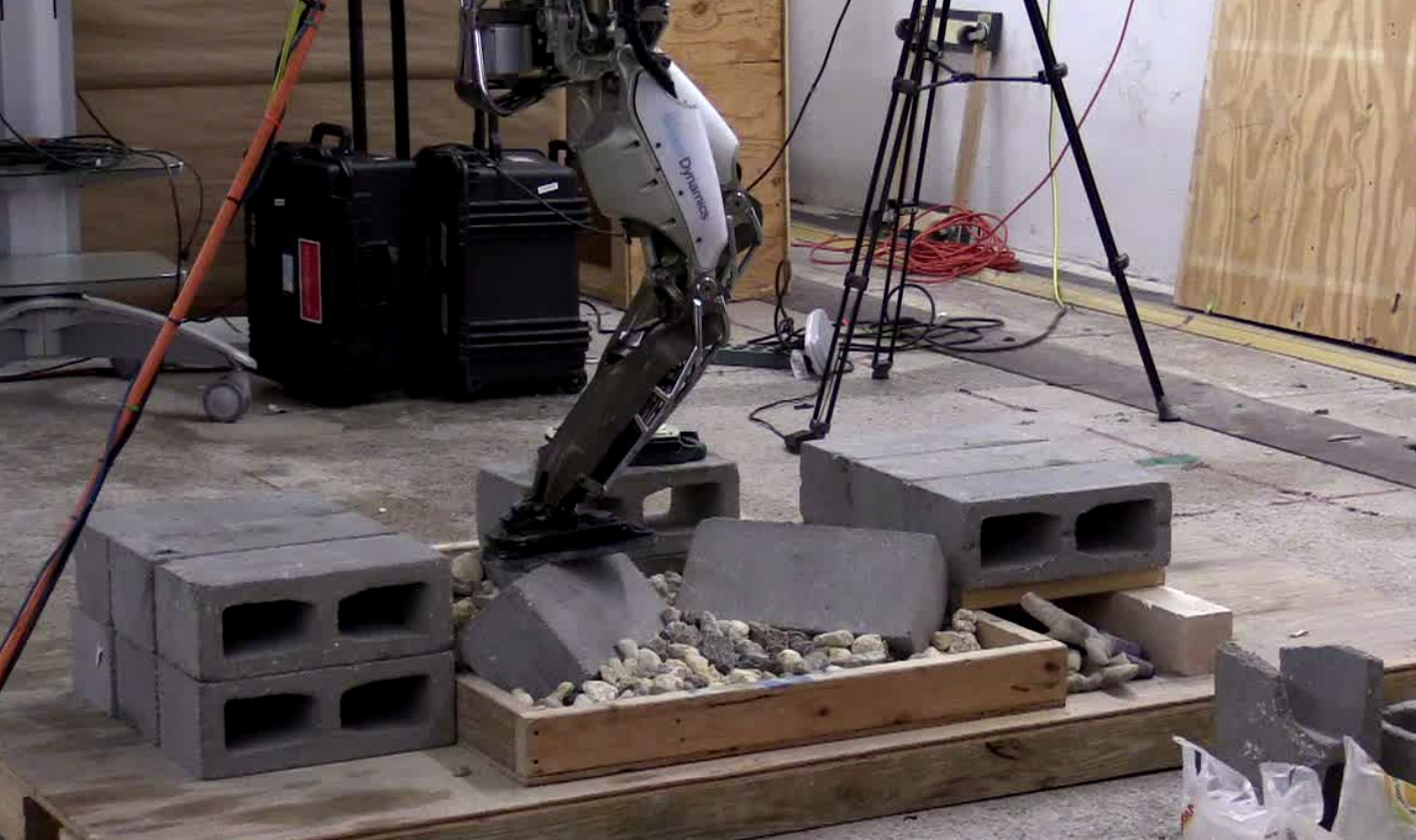}
  }\vspace{0.1cm}
  \raisebox{-0.5\height}{%
    \includegraphics[width=0.85\columnwidth]{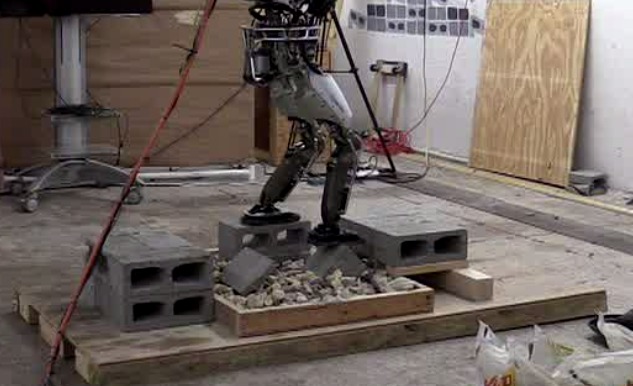}
  }
  \caption{Atlas walking over variously oriented line contacts. We use tilted cinder blocks to produce line contacts. It can be seen that the feet of the robot stay flat and do not conform to the cinder block surfaces.}
  \label{fig:line_testbed}
\end{figure}

\begin{figure*}
  \centering
  \includegraphics[width=0.77\textwidth]{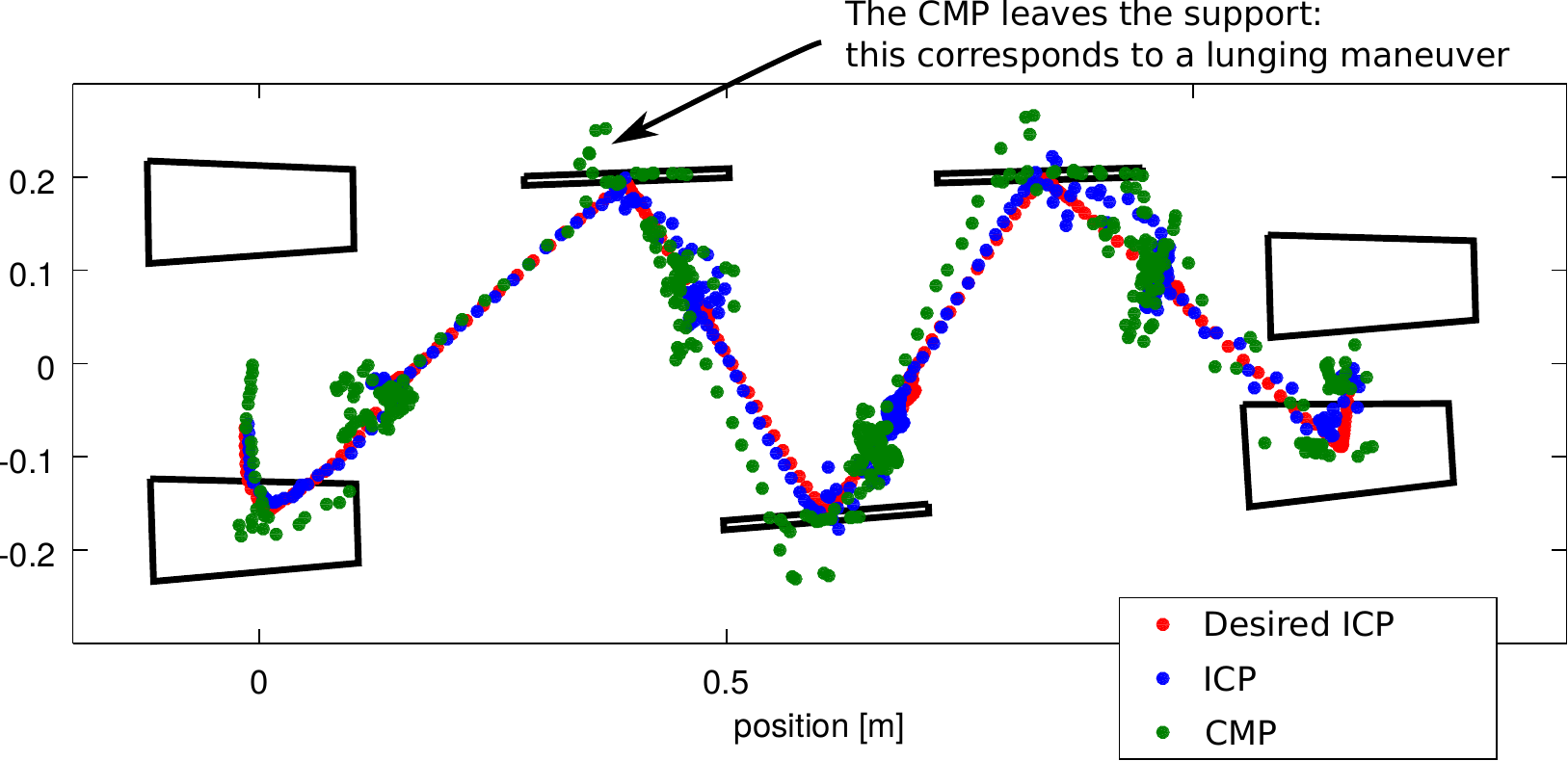}
  \caption{This plot shows a top view of the measured ground reference points during a run with the real Atlas robot (walking from left to right). The robot walks over the straight cinder block field seen in Fig. \ref{fig:straight_lines}. The black polygons show the detected footholds after line fitting. Every tenth data point is shown, each a duration of 10\,ms seconds apart.}
  \label{fig:reference_point_plot}
\end{figure*}

\begin{figure*}
  \centering
  \raisebox{-0.5\height}{%
    \includegraphics[height=3.7cm]{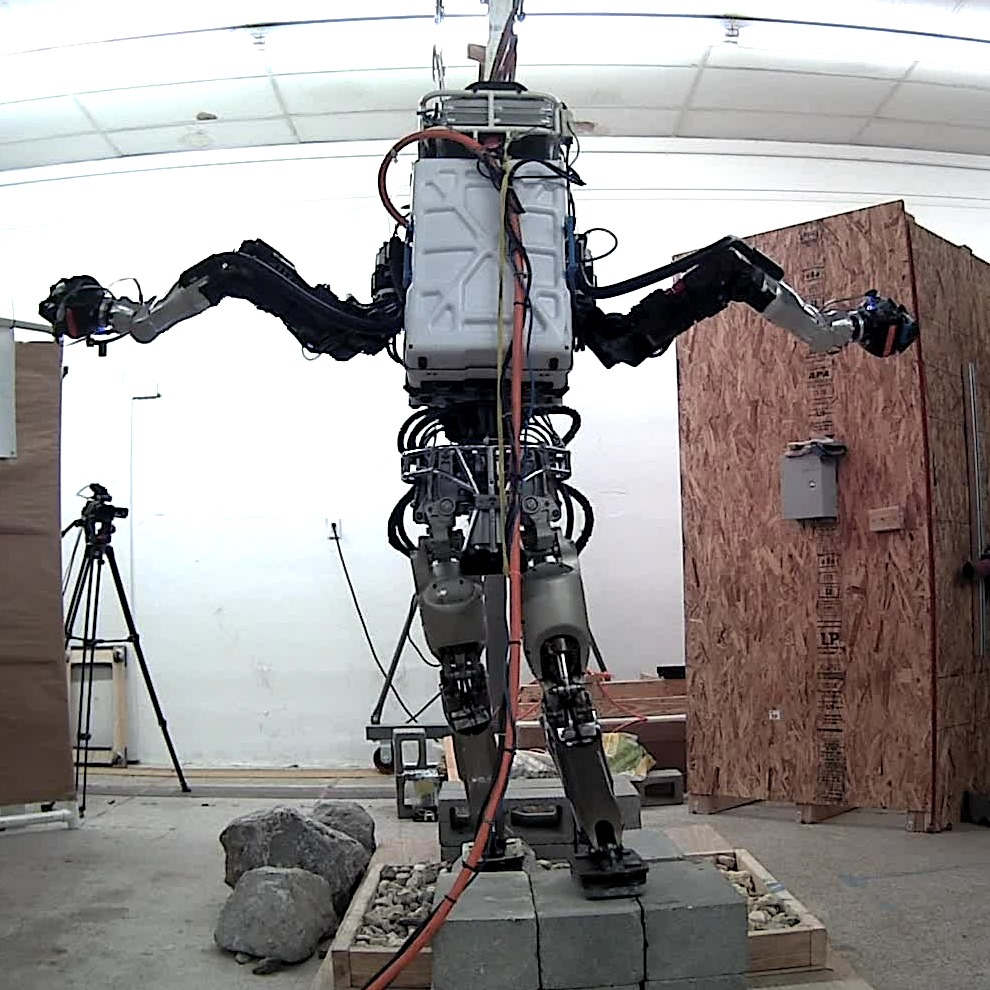}
  }\vspace{0.03cm}
  \raisebox{-0.5\height}{%
    \includegraphics[height=3.7cm]{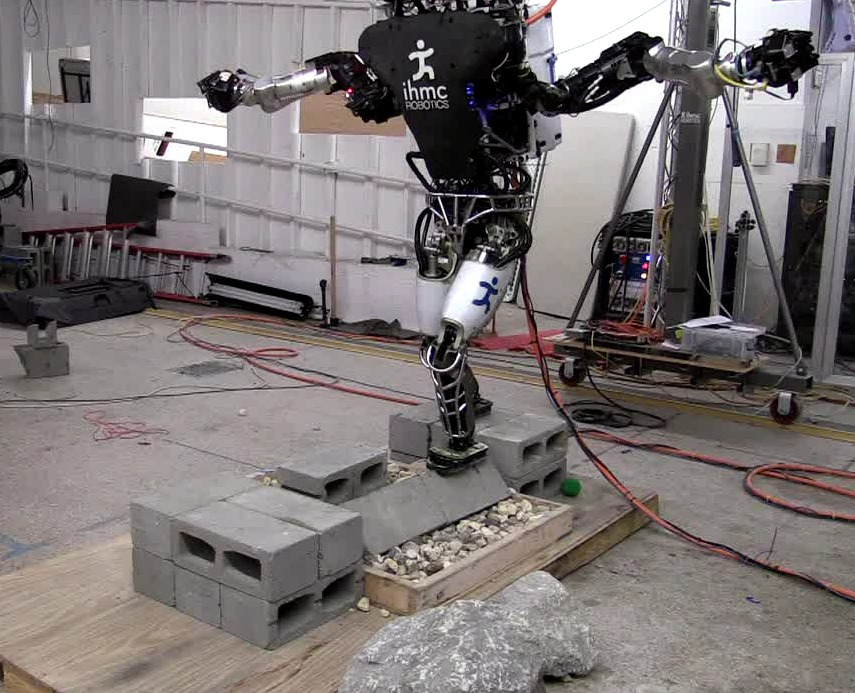}
  }
  \raisebox{-0.5\height}{%
    \includegraphics[height=3.7cm]{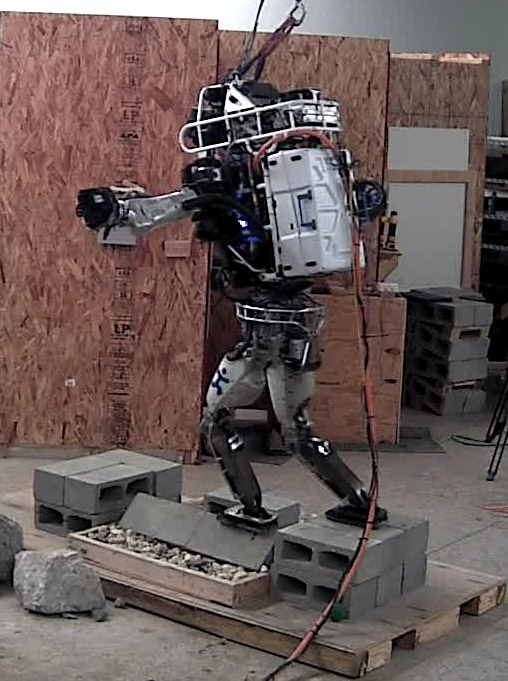}
  }\vspace{0.03cm}
  \raisebox{-0.5\height}{%
    \includegraphics[height=3.7cm]{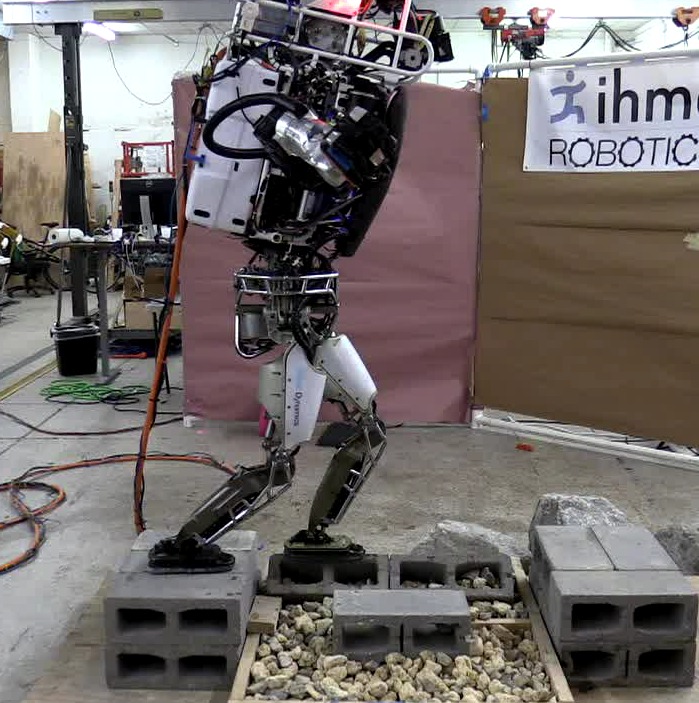}
  }\\
  \raisebox{-0.5\height}{%
    \includegraphics[height=3.7cm]{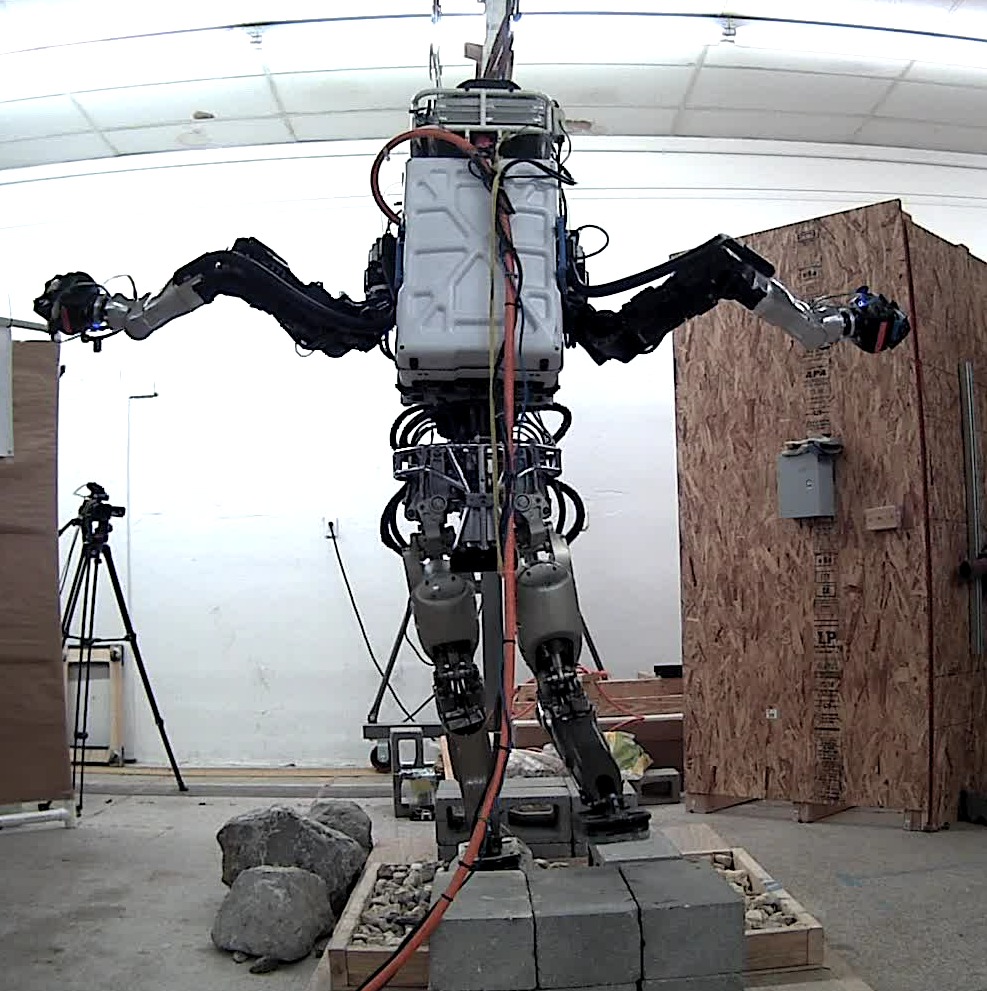}
  }\vspace{0.03cm}
  \raisebox{-0.5\height}{%
    \includegraphics[height=3.7cm]{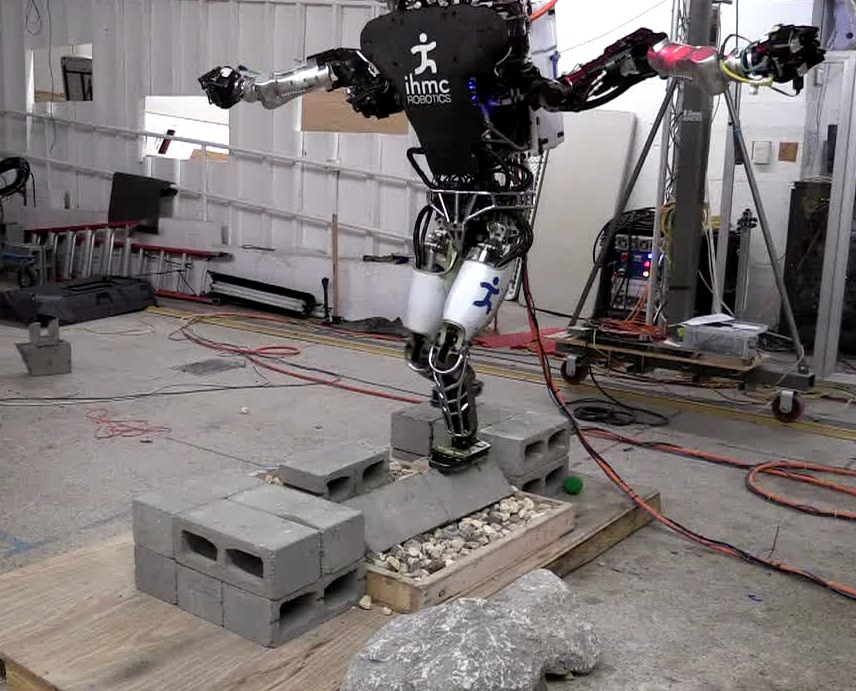}
  }
  \raisebox{-0.5\height}{%
    \includegraphics[height=3.7cm]{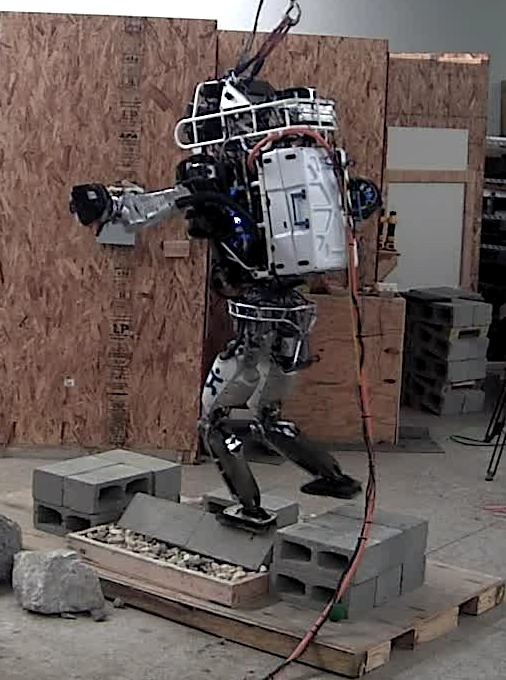}
  }\vspace{0.03cm}
  \raisebox{-0.5\height}{%
    \includegraphics[height=3.7cm]{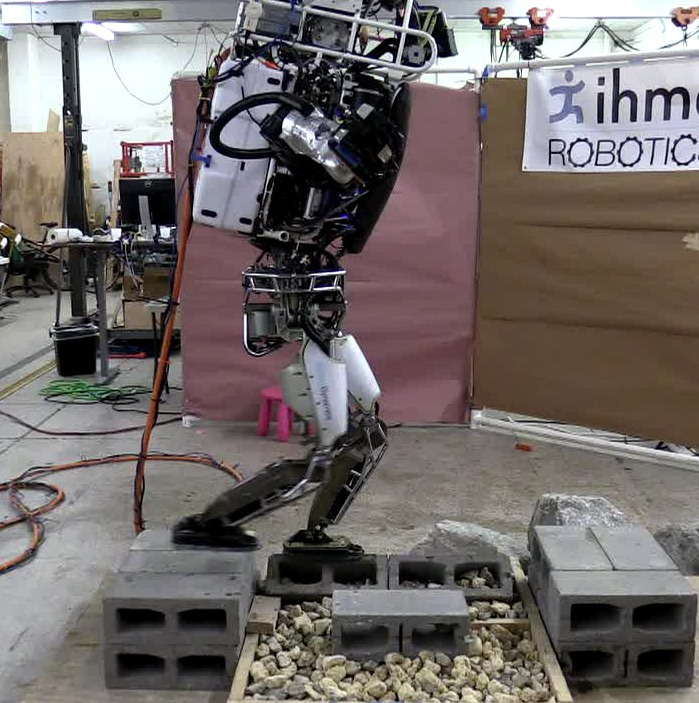}
  }\\
  \raisebox{-0.5\height}{%
    \includegraphics[height=3.7cm]{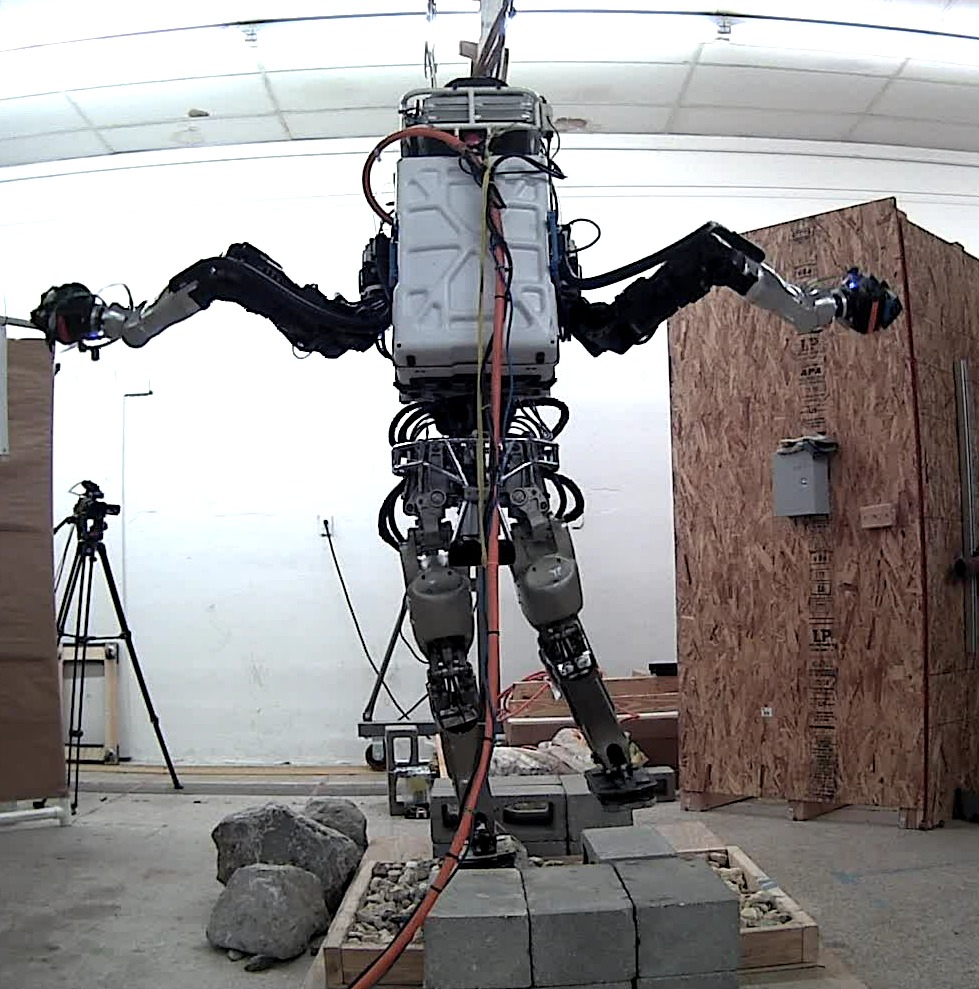}
  }\vspace{0.03cm}
  \raisebox{-0.5\height}{%
    \includegraphics[height=3.7cm]{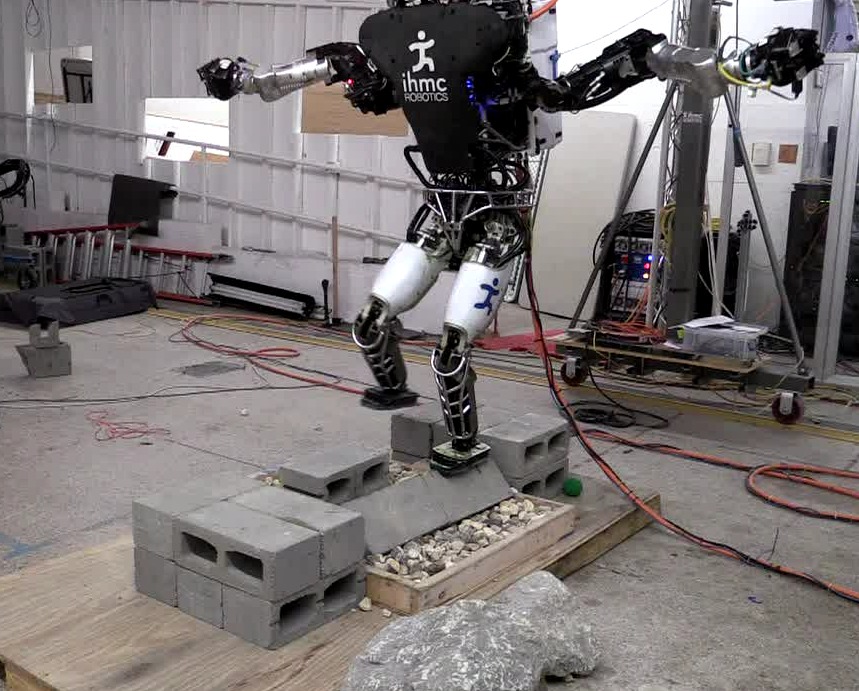}
  }
  \raisebox{-0.5\height}{%
    \includegraphics[height=3.7cm]{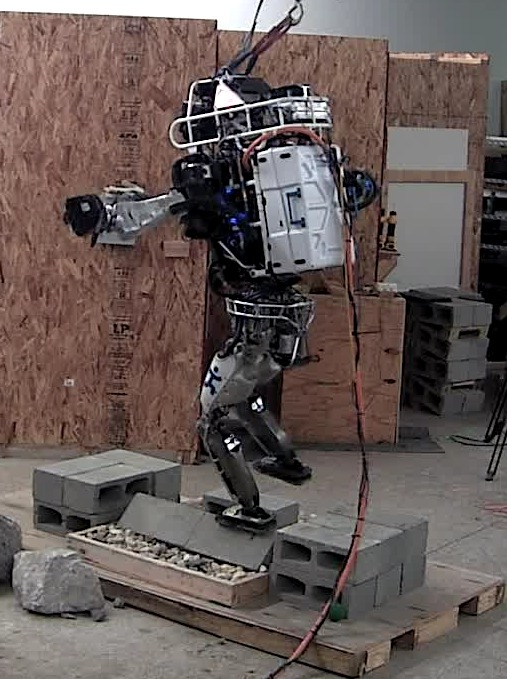}
  }\vspace{0.03cm}
  \raisebox{-0.5\height}{%
    \includegraphics[height=3.7cm]{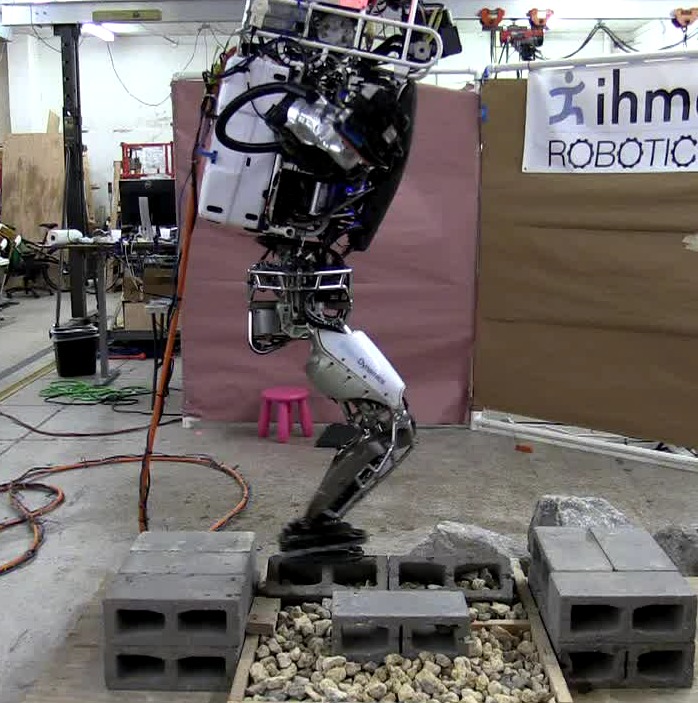}
  }\\
  \raisebox{-0.5\height}{%
    \includegraphics[height=3.7cm]{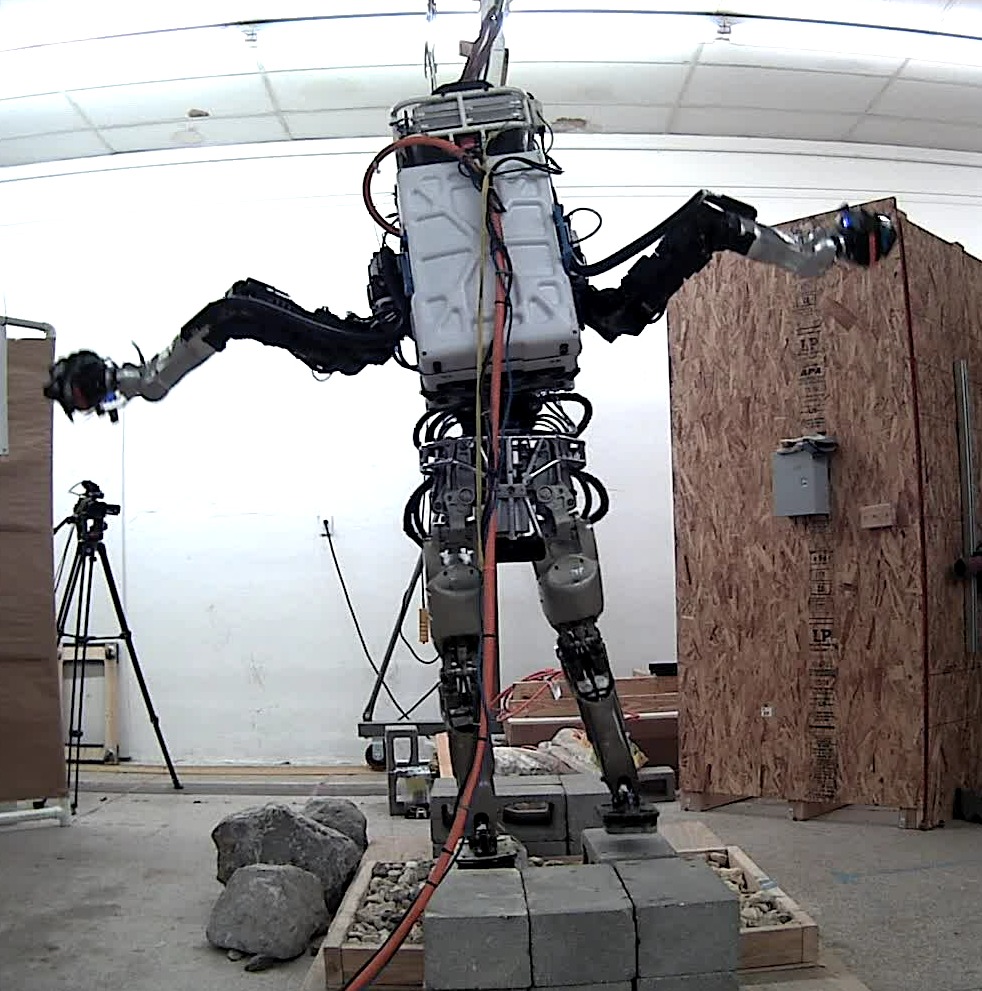}
  }\vspace{0.03cm}
  \raisebox{-0.5\height}{%
    \includegraphics[height=3.7cm]{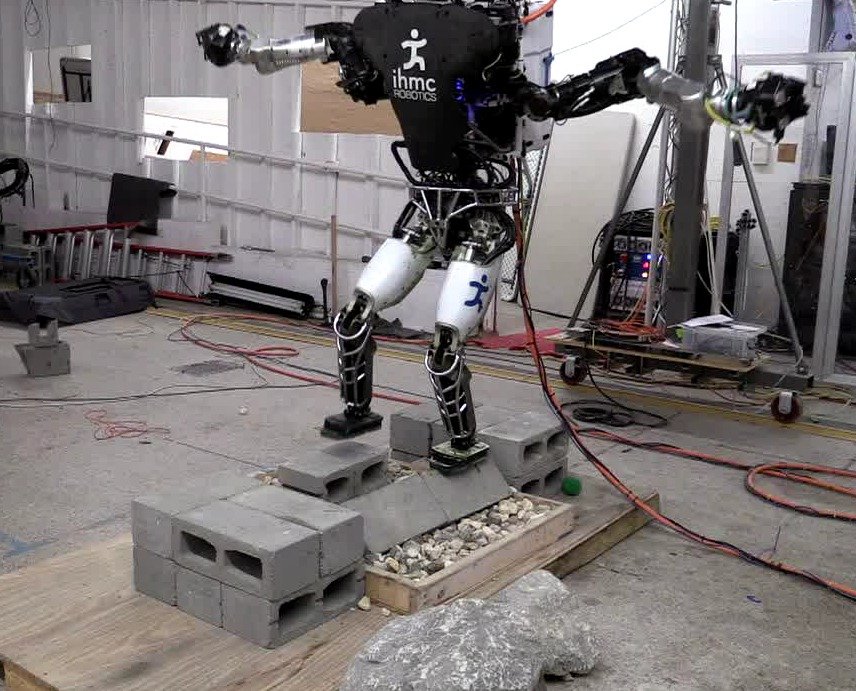}
  }
  \raisebox{-0.5\height}{%
    \includegraphics[height=3.7cm]{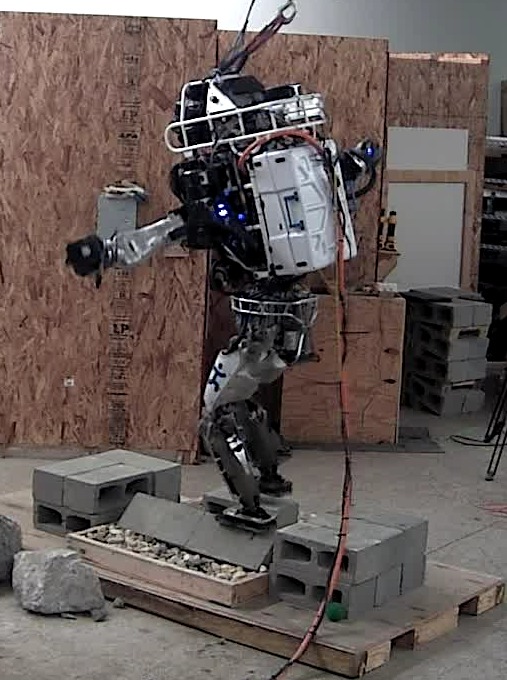}
  }\vspace{0.03cm}
  \raisebox{-0.5\height}{%
    \includegraphics[height=3.7cm]{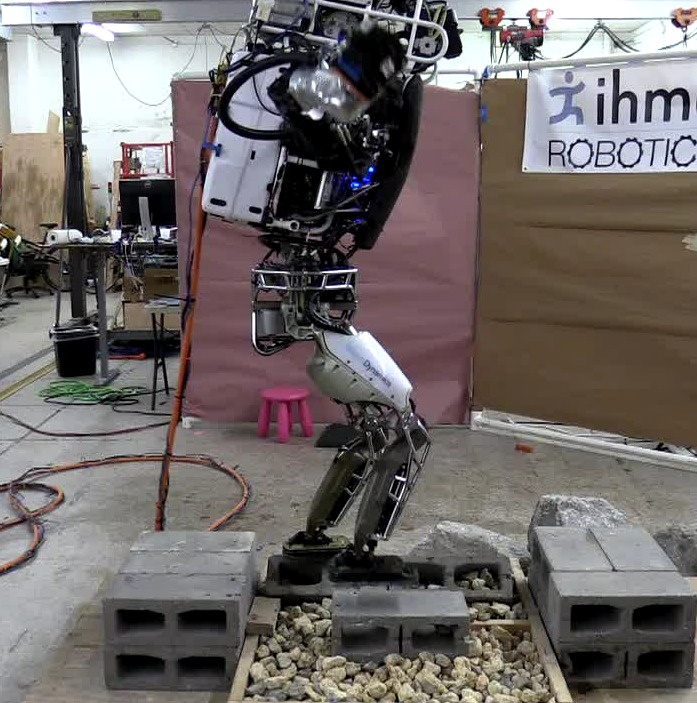}
  }\\
  \raisebox{-0.5\height}{%
    \includegraphics[height=3.7cm]{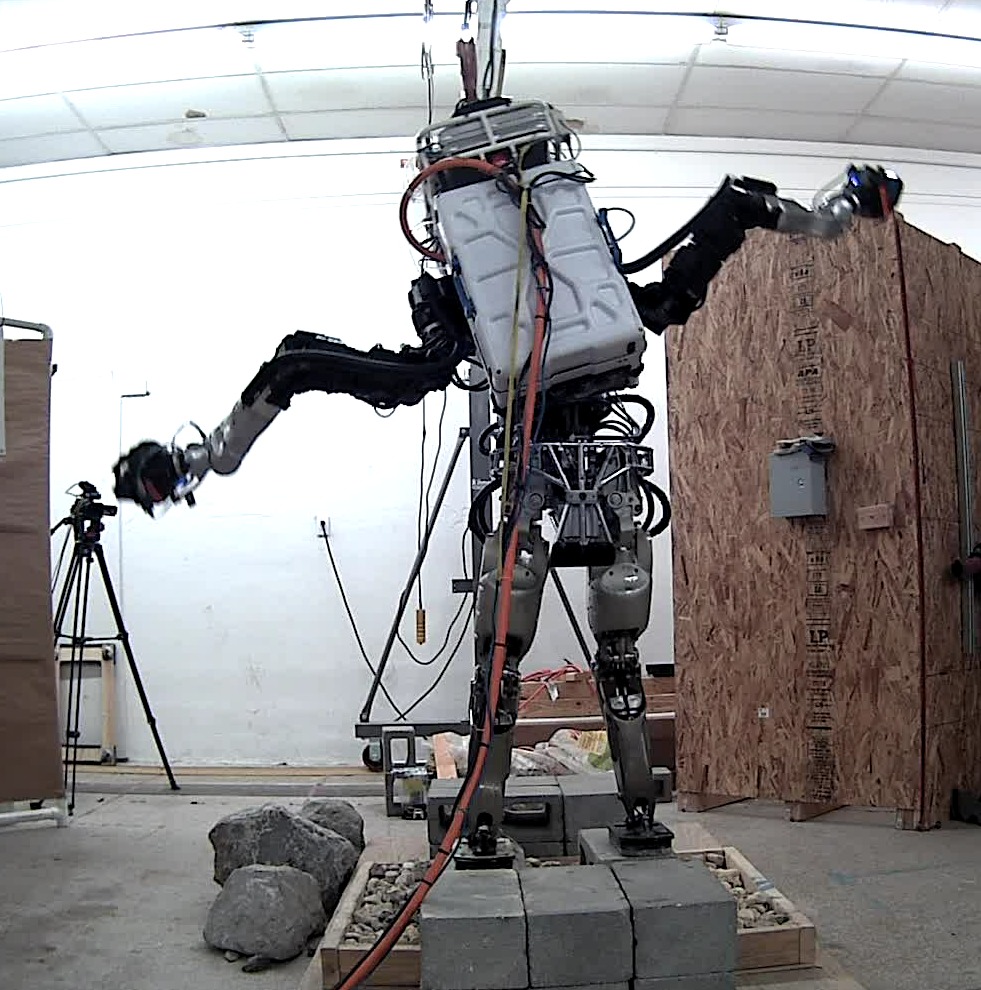}
  }\vspace{0.03cm}
  \raisebox{-0.5\height}{%
    \includegraphics[height=3.7cm]{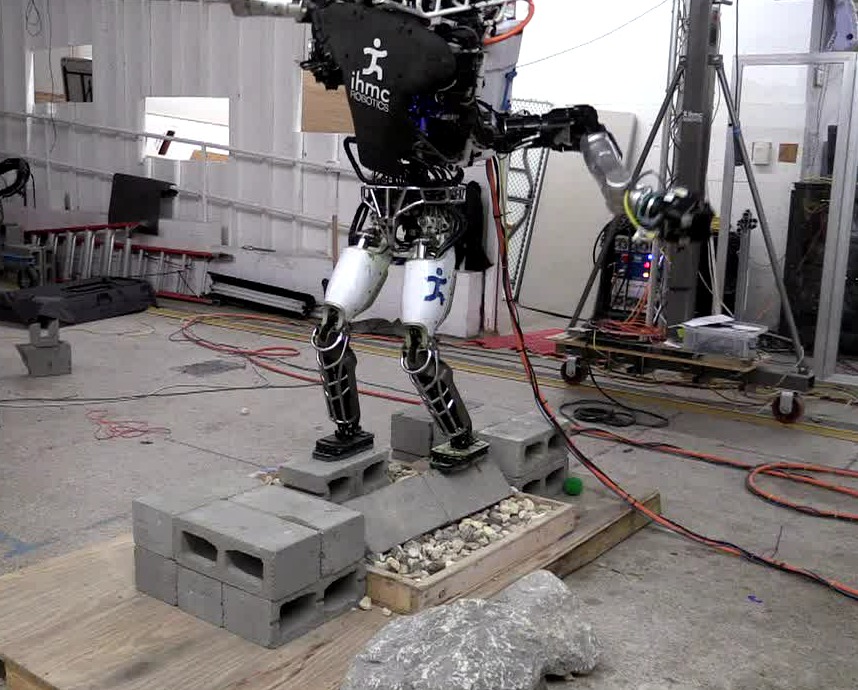}
  }
  \raisebox{-0.5\height}{%
    \includegraphics[height=3.7cm]{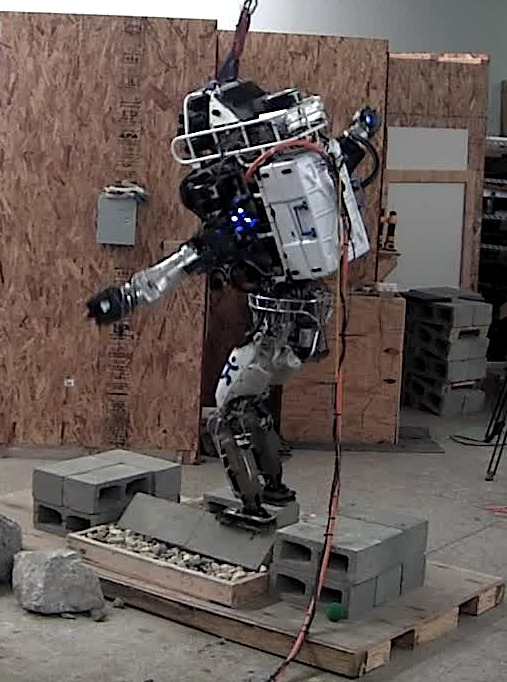}
  }\vspace{0.03cm}
  \raisebox{-0.5\height}{%
    \includegraphics[height=3.7cm]{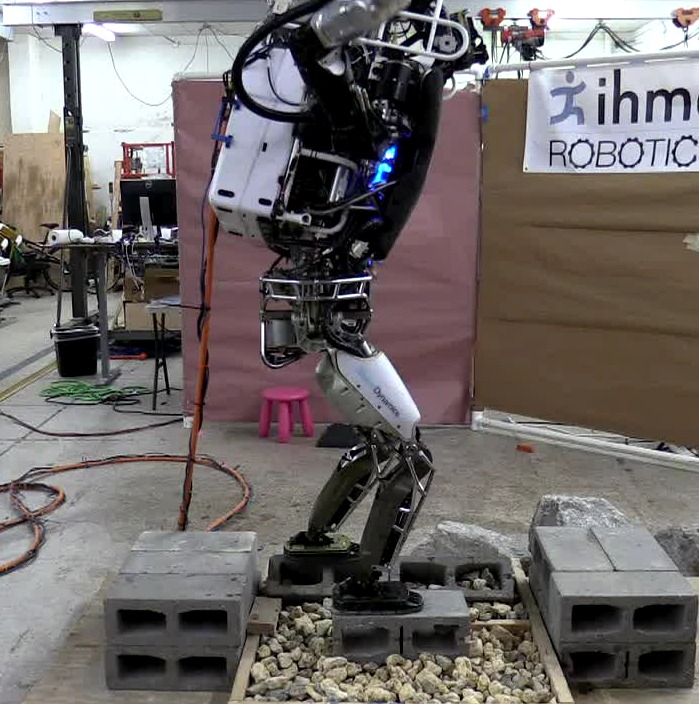}
  }\\
  \raisebox{-0.5\height}{%
    \includegraphics[height=3.7cm]{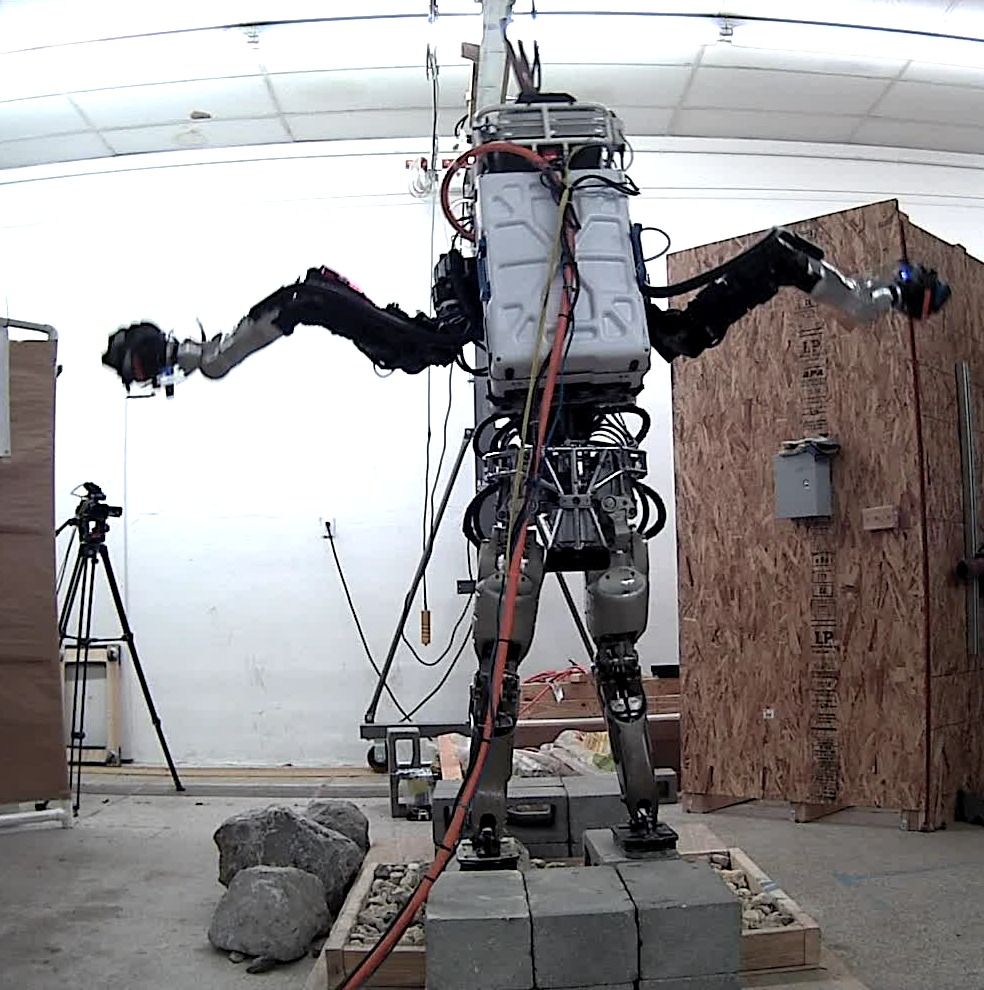}
  }\vspace{0.03cm}
  \raisebox{-0.5\height}{%
    \includegraphics[height=3.7cm]{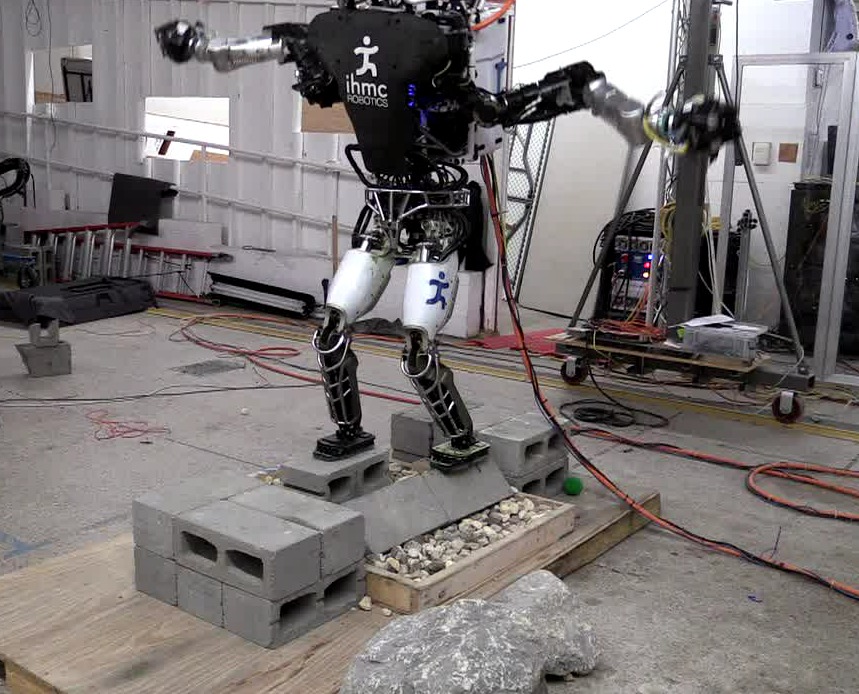}
  }
  \raisebox{-0.5\height}{%
    \includegraphics[height=3.7cm]{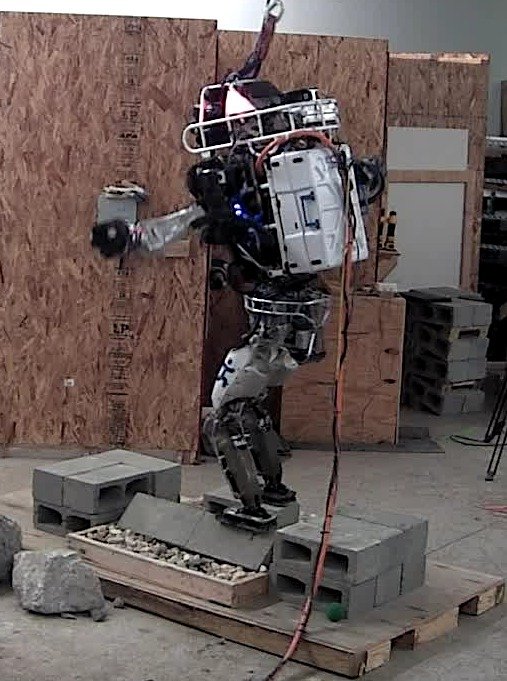}
  }\vspace{0.03cm}
  \raisebox{-0.5\height}{%
    \includegraphics[height=3.7cm]{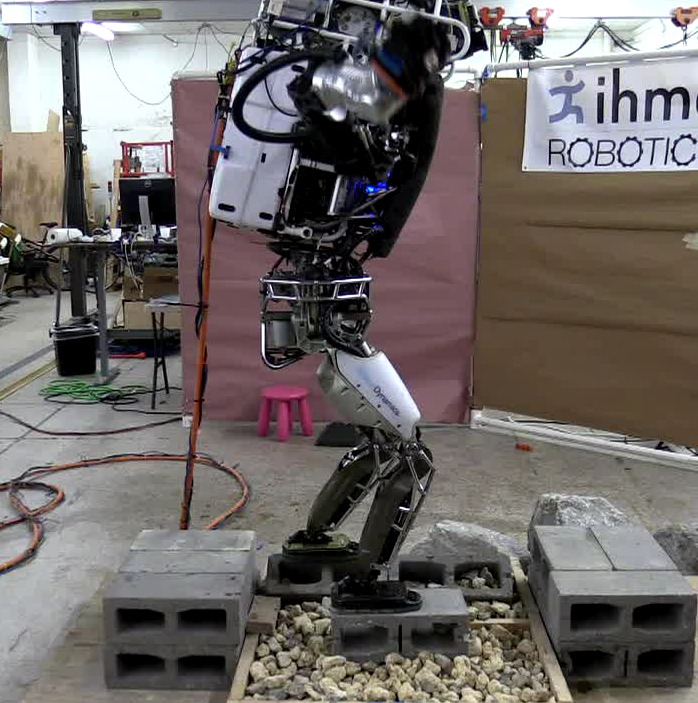}
  }\\
  \caption{Picture sequence showing the Atlas robot taking a step with the stance foot only partially supported. The pictures in one row are taken at the same time. Each column shows the robot stepping from a different camera angle. During the step the robot lunges its upper body to the side to prevent a fall to the left. After the step is completed the upper body is brought back into a upright pose.}
  \label{fig:step_picture_sequence}
\end{figure*}

The real robot was able to step over single, randomly oriented lines (Fig. \ref{fig:line_testbed}). We created line contacts by tilting cinder blocks to a 45$^\circ$ angle. In the case of straight, forward oriented lines the real robot was able to continuously walk on the line contacts for several steps in a row (Fig. \ref{fig:straight_lines}). The incorporation of angular momentum to maintain balance produces human like balancing motions. In Fig. \ref{fig:reference_point_plot}, a top view of the ground reference point trajectories is shown. They were recorded during a run with the real Atlas robot walking over line contacts. It can be seen that for the line footholds the CMP leaves the support area towards the outside to prevent the robot from falling and producing angular momentum. Finally, Fig. \ref{fig:step_picture_sequence} shows a sequence of pictures taken during a step with the stance foot on a angled line. In this sequence a sideways lunging motion can be seen that helps the robot maintain balance and finish the step.

Our algorithm requires adequate control of the CoP and CMP. On the Atlas robot, the CoP is controllable with an accuracy of approximately 2\,cm due to good force control in the ankle joints. This is verified using the foot force sensors. To improve CMP tracking we add a joint velocity control term based on integrated desired accelerations to the leg and back joints [12]. Since there is no sensor for measuring the CMP directly, it is difficult to quantify how well the robot is able to control it.

As long as the combination of left and right footholds provides a sufficiently large support area we observed that the robot was able to regain balance after each step. If the footholds are alternating between full and partial, the simulated robot is able to explore and walk over a point stepping stone (actual size 2\,cm$\times$2\,cm). In reality the performance of the robot was limited by multiple factors:

\begin{itemize}
 \item The torque and velocity limits of the actuators on the robot bound the stepping speed and the change in angular momentum achievable. As hardware evolves, we are expecting to see robots perform better and better at tasks like balancing that reach the physical limits of the robot.
 \item Maximum joint angles of the hip and spine joints cap the amount of lunging motion. This is expected and similar limits are found in humans. The ICP algorithm we use for balancing relies on the LIPM with flywheel as model. It does not incorporate knowledge of the joint limits. To improve our algorithms we are working on extending our controller to incorporate knowledge of range of motion and other limitations.
 \item Finally, noisy sensors prevent precise ICP calculations and hinder the foothold estimation. Our simulations include sensor noise but disturbances, elasticities, and inaccuracies on the real robot make balancing on small footholds a hard problem. Velocity estimations are usually particularly noisy or, if filtered, delayed. We aim to improve our state estimator to provide fast and precise robot state measurements. A promising approach on improving joint velocity measurements using link mounted IMUs has been proposed in \cite{xinjilefu2016distributed}.
\end{itemize}


\section{CONCLUSION}

Using the presented walking framework we were able to walk over partial footholds such as line contacts with the Atlas humanoid. This is an important step in the effort of making legged robots useful in real world scenarios. The ability to walk on unexpected partial footholds greatly increases the robustness of a robot when employed in cluttered environments. In addition it vastly extends the set of environments a robot can traverse. We use angular momentum to regain balance when other strategies cannot prevent a fall. In the future, we hope to improve the balancing capabilities of our robot by improving our control algorithms and state estimation and by applying the algorithms to robots with higher joint velocity limits.


\bibliographystyle{bib/IEEEtran}
\bibliography{./bib/IEEEabrv.bib,./bib/myBib.bib}

\end{document}